\documentclass[sigconf,language=english]{acmart}

\usepackage{hyperref}
\usepackage{url}
\usepackage{xspace}
\usepackage{pgfplots}
\pgfplotsset{width=13cm, compat=1.6, log ticks with fixed point}
\usetikzlibrary{patterns}
\usetikzlibrary{arrows}
\usepackage[algoruled, vlined, procnumbered]{algorithm2e}

\newcommand{\previoustool}{\textsc{DiskANN}\xspace}
\newcommand{\tool}{\textsc{OOD-DiskANN}\xspace}

\newcommand{\faketool}{\textsc{SCANN}\xspace}
\newcommand{\gorder}{\textsc{Gorder}\xspace}
\newcommand{\parallelgorder}{\textsc{ParallelGorder}\xspace}
\newcommand{\faiss}{\textsc{FAISS-IVF}\xspace}
\newcommand{\faisspq}{\textsc{FAISS-IVFPQ}\xspace}
\newcommand{\hnsw}{\textsc{HNSW}\xspace}
\newcommand{\nsg}{\textsc{NSG}\xspace}
\newcommand{\lsh}{\textsc{Cross Polytope LSH}\xspace}
\newcommand{\vamana}{\textsc{Vamana}\xspace}
\newcommand{\greedysearch}{\textsc{GreedySearch}\xspace}
\newcommand{\robustprune}{\textsc{RobustPrune}\xspace}
\newcommand{\robuststitch}{\textsc{RobustStitch}\xspace}
\newcommand{\robustvamana}{\textsc{RobustVamana}\xspace}

\newcommand{\sectorpack}{\textsc{SectorPack}\xspace}

\newcommand\mydots{\hbox to 1em{.\hss.\hss.}}

\newcommand{\recall}[1]{\textit{#1}\text{-recall@}\textit{#1}}

\mathchardef\mhyphen="2D

\definecolor{bblue}{HTML}{4F81BD}
\definecolor{rred}{HTML}{C0504D}
\definecolor{ggreen}{HTML}{9BBB59}
\definecolor{ppurple}{HTML}{8A2BE2} 
\definecolor{yyellow}{HTML}{DDAE06}

\tikzstyle{bar1} = [bblue, fill=bblue, postaction={pattern=north east lines}]
\tikzstyle{bar2} = [rred, fill=rred]
\tikzstyle{bar3} = [ggreen, fill=ggreen, postaction={pattern=north west lines}]
\tikzstyle{bar4} = [ppurple, fill=ppurple, postaction={pattern=crosshatch dots}]
\tikzstyle{bar5} = [yyellow, fill=yyellow, postaction={pattern=grid}]

\let\oldnl\nl
\newcommand{\nonl}{\renewcommand{\nl}{\let\nl\oldnl}}

\usepackage{lipsum,adjustbox}
\usepackage[english]{babel}
\addto\extrasenglish{

}

\AtBeginDocument{%
  \providecommand\BibTeX{{%
    \normalfont B\kern-0.5em{\scshape i\kern-0.25em b}\kern-0.8em\TeX}}}

\setcopyright{none}
\copyrightyear{2023}
\acmYear{2023}
\acmDOI{XXXXXXX.XXXXXXX}

\acmConference[Under Review]{The ACM Web Conference
2023}{April 30--May 04,
  2023}{Austin, TX}

\begin{document}

\title{OOD-DiskANN: Efficient and Scalable Graph ANNS for Out-of-Distribution Queries}

\author{Shikhar Jaiswal}
\authornote{Work done during Research Fellowship at MSR India.}
\email{t-sjaiswal@microsoft.com}
\affiliation{
  \institution{Microsoft Research India}
  \country{India}
}

\author{Ravishankar Krishnaswamy}
\email{rakri@microsoft.com}
\affiliation{
  \institution{Microsoft Research India}
  \country{India}
}

\author{Ankit Garg}
\email{garga@microsoft.com}
\affiliation{
  \institution{Microsoft Research India}
  \country{India}
}

\author{Harsha Vardhan Simhadri}
\email{harshasi@microsoft.com}
\affiliation{
  \institution{Microsoft Research India}
  \country{India}
}

\author{Sheshansh Agrawal}
\email{sheshansh.agrawal@microsoft.com}
\affiliation{
  \institution{Microsoft}
  \country{USA}
}


\begin{abstract}
State-of-the-art algorithms for Approximate Nearest Neighbor Search (ANNS) such as \previoustool, \faiss, and \hnsw build data dependent indices that offer substantially better accuracy and search efficiency over data-agnostic indices by overfitting to the index data distribution. 
When the query data is drawn from a different distribution -- e.g., when index represents image embeddings and query represents textual embeddings -- such algorithms lose much of this performance advantage.
On a variety of datasets, for a fixed recall target, latency is worse by an order of magnitude or more for Out-Of-Distribution (OOD) queries as compared to In-Distribution (ID) queries.
The question we address in this work is whether ANNS algorithms can be made efficient for OOD queries if the index construction is given access to a small sample set of these queries.
We answer positively by presenting \tool, which uses a sparing sample ($1\%$ of index set size) of OOD queries, and provides up to $40\%$ improvement in mean query latency over SoTA algorithms of a similar memory footprint.
\tool is scalable and has the efficiency of graph-based ANNS indices.
Some of our contributions can improve query efficiency for ID queries as well.
\end{abstract}

\ccsdesc[500]{Theory of computation~Data compression}
\ccsdesc[500]{Theory of computation~Nearest neighbor algorithms}
\ccsdesc[300]{Information systems~Retrieval models and ranking}

\keywords{Approximate Nearest Neighbor Search (ANNS), Query-Aware Product Quantization (Query-Aware PQ), Graph Algorithms}



\maketitle

\section{Introduction}
\label{sec:introduction}


Embedding-based retrieval (aka semantic- or dense-retrieval) is increasingly the paradigm of choice for search and recommendation systems across various domains such as document retrieval~\citep{DPR20}, web relevance~\citep{ANCE21,TREC-DL19,Semantic-models-IR-survey22}, advertisement~\citep{XC} and content-based image retrieval~\citep{long2003fundamentals,wan2014deep} to name a few. These systems rely on searching an index built over the embeddings corresponding to the objects of interest, to retrieve the nearest embeddings to a query's embedding, based on some geometric distance (such as $\ell_2$).
Since solving the problem exactly requires an expensive exhaustive scan of the database -- which would be impractical for real-world indices that span billions of objects --  practical interactive search systems use Approximate Nearest Neighbor Search (ANNS) algorithms with highly sub-linear query complexity~\citep{LSH, IVF-OADCGP, DiskANN, HNSW} to answer such queries. The quality of such ANN indices is often measured by \recall{k} which is the overlap between the top-$k$ results of the index search with the ground truth $k$-nearest neighbors ($k$-NNs) in the corpus for the query, averaged over a representative query set. 

State-of-the-art algorithms for ANNS, such as graph-based indices~\citep{HNSW, NSG, DiskANN} which use data-dependent index construction, achieve better query efficiency over prior data-agnostic methods like LSH~\citep{FALCONN, LSH} (see \textbf{\autoref{sec:anns-preliminaries}} for more details). 
Such efficiency enables these indices to serve queries with $>90\%$ recall with a latency of a few milliseconds, required in interactive web scenarios. They do so by building an index that makes it easy to query a point from the indexed (also called base) dataset itself, thereby enabling the index to answer queries that are drawn roughly from the same distribution. Embedding-based retrieval, via such ANNS indices, is widely deployed for web services in the industry~\citep{SimSearchNet-v2}.

With the advancement of multi-modal embeddings~\citep{CLIP, YandexT2I, T-Bletchley}, there is an increasing need for accurate ANNS algorithms that work well with ``cross-modal queries''. As a motivating example, consider a situation where a user searches through an image index with only a textual description as input. Jointly learnt image-text models can preserve query semantics well, but the text embeddings they generate often lie in a different distribution than the image embeddings, even if both the embeddings share the same representation space. 

\begin{figure*}[t]
\flushleft
\begin{minipage}[c]{\columnwidth}
    \begin{tikzpicture}[scale=0.6,font=\large]
    \begin{semilogyaxis}[ 
        title=FAISS-IVF,
        width=\linewidth,
        line width=0.8,
        grid=both,
        grid style={line width=.1pt, draw=gray!10},
        major grid style={line width=.2pt,draw=gray!50},
        minor grid style={line width=.3pt,draw=gray!100, dotted},
        minor tick num=3,
        tick label style={font=\large},
        legend style={nodes={scale=1.0, transform shape}},
        label style={font=\large},
        grid style={white},
        xlabel={10-Recall@10},
        ylabel={Mean Query Latency (ms)},
        legend style={at={(0,-0), font=\normalsize}, anchor=south west, draw=none, fill=none},
        legend columns=2,
        ymin=0.1,
        ymax=100,
        xmin=10,
        xmax=100,
    ]
        \addplot[bblue, mark=square*] coordinates
          {(76.15,0.698) (85.33,1.054) (89.11,1.531) (91.28,1.887) (92.73,2.359) (93.76,2.894) (94.56,3.207) (95.18,3.609) (95.67,4.035) (96.08,4.372)};
          \addlegendentry{Web Ads ID}

        \addplot[rred, mark=square*] coordinates
          {(74.44,0.594) (83.87,0.905) (87.73,1.263) (89.93,1.709) (91.40,2.138) (92.46,2.378) (93.27,2.690) (93.92,3.077) (94.45,3.480) (94.90,3.754)};
          \addlegendentry{Web Ads OOD}

        \addplot[bblue, mark=otimes*] coordinates
          {(75.86,2.080) (87.56,3.538) (92.00,5.126) (94.29,6.713) (95.65,8.482) (96.54,10.072) (97.17,11.640) (97.61,13.098) (97.96,14.567) (98.23,16.457)};
          \addlegendentry{Yandex T2I ID}

        \addplot[rred, mark=otimes*] coordinates
          {(41.95,1.760) (54.69,2.814) (61.73,4.159) (66.42,5.549) (69.84,6.961) (72.50,8.448) (74.63,9.673) (76.38,10.883) (77.88,12.215) (79.14,13.433)};
          \addlegendentry{Yandex T2I OOD}

        \addplot[bblue, mark=triangle*] coordinates
          {(70.56,12.901) (82.46,19.832) (87.51,27.460) (90.33,36.148) (92.13,44.709) (93.38,53.605) (94.31,62.410) (95.00,74.112) (95.58,81.869) (96.03,89.791)};
          \addlegendentry{Turing T2I ID}

        \addplot[rred, mark=triangle*] coordinates
          {(13.14,8.980) (20.27,15.351) (25.20,21.692) (28.77,28.988) (31.58,36.298) (33.98,43.797) (36.02,50.569) (37.78,58.062) (39.41,65.615) (40.81,72.814)};
          \addlegendentry{Turing T2I OOD}
    \end{semilogyaxis}
    \hspace{4.7cm}
    \begin{semilogyaxis}[ 
        title=HNSW,
        width=\linewidth,
        line width=0.8,
        grid=both,
        grid style={line width=.1pt, draw=gray!10},
        major grid style={line width=.2pt,draw=gray!50},
        minor grid style={line width=.3pt,draw=gray!100, dotted},
        minor tick num=3,
        tick label style={font=\large},
        legend style={nodes={scale=1.0, transform shape}},
        label style={font=\large},
        legend image post style={mark=triangle*},
        grid style={white},
        xlabel={10-Recall@10},
        legend style={at={(0,1), font=\normalsize}, anchor=north west, draw=none, fill=none},
        legend columns=1,
        ymin=0.08,
        ymax=10,
        xmin=40,
        xmax=100,
    ]
        \addplot[bblue, mark=square*] coordinates
          {(83.49,0.097) (90.58,0.137) (93.41,0.195) (94.95,0.259) (95.95,0.317) (96.63,0.343) (97.12,0.349) (97.50,0.403) (97.80,0.464) (98.04,0.467)};

        \addplot[rred, mark=square*] coordinates
          {(78.42,0.100) (87.37,0.140) (90.99,0.203) (92.94,0.267) (94.22,0.311) (95.12,0.335) (95.80,0.389) (96.30,0.405) (96.70,0.414) (97.04,0.474)};

        \addplot[bblue, mark=otimes*] coordinates
          {(82.15,0.322) (91.16,0.500) (94.47,0.535) (96.09,0.555) (97.07,0.636) (97.70,0.722) (98.11,0.825) (98.42,0.959) (98.65,1.056) (98.82,1.123)};

        \addplot[rred, mark=otimes*] coordinates
          {(43.55,0.345) (56.97,0.551) (64.22,0.599) (68.98,0.658) (72.39,0.783) (74.98,0.881) (77.04,1.008) (78.70,1.131) (80.10,1.159) (81.26,1.349)};

        \addplot[bblue, mark=triangle*] coordinates
          {(85.57,1.099) (93.07,1.755) (95.69,2.335) (97.01,2.427) (97.77,2.525) (98.29,2.614) (98.60,2.831) (98.82,3.056) (99.00,3.267) (99.13,3.389)};

        \addplot[rred, mark=triangle*] coordinates
          {(42.75,1.161) (56.29,1.982) (63.48,2.003) (68.13,2.856) (71.48,2.906) (74.02,3.176) (76.04,3.513) (77.70,3.585) (79.09,3.811) (80.25,4.731)};
    \end{semilogyaxis}
    \hspace{4.7cm}
    \begin{semilogyaxis}[ 
        title= Vamana (DiskANN),
        width=\linewidth,
        line width=0.8,
        grid=both,
        grid style={line width=.1pt, draw=gray!10},
        major grid style={line width=.2pt,draw=gray!50},
        minor grid style={line width=.3pt,draw=gray!100, dotted},
        minor tick num=3,
        tick label style={font=\large},
        legend style={nodes={scale=1.0, transform shape}},
        label style={font=\large},
        grid style={white},
        xlabel={10-Recall@10},
        legend style={at={(1,1), font=\normalsize}, anchor=north west, draw=none, fill=none},
        legend columns=1,
        ymin=0.08,
        ymax=10,
        xmin=50,
        xmax=100,
    ]
        \addplot[bblue, mark=square*] coordinates
          {(88.03,0.147) (93.47,0.199) (95.50,0.204) (96.61,0.251) (97.29,0.299) (97.77,0.352) (98.13,0.370) (98.40,0.410) (98.62,0.464) (98.77,0.501)};

        \addplot[rred, mark=square*] coordinates
          {(83.96,0.151) (91.06,0.203) (93.70,0.212) (95.12,0.262) (96.02,0.300) (96.68,0.347) (97.16,0.401) (97.52,0.419) (97.77, 0.449) (97.99,0.485)};

        \addplot[bblue, mark=otimes*] coordinates
          {(90.90,0.412) (96.37,0.419) (98.02,0.535) (98.76,0.623) (99.13,0.742) (99.35,0.850) (99.48,0.944) (99.58,1.079) (99.66,1.122) (99.70,1.279)};

        \addplot[rred, mark=otimes*] coordinates
          {(54.19,0.450) (67.96,0.475) (74.73,0.617) (79.00,0.701) (81.77,0.807) (83.79,0.902) (85.41,1.019) (86.69,1.148) (87.80,1.296) (88.67,1.455)};

        \addplot[bblue, mark=triangle*] coordinates
          {(91.20,1.463) (96.09,1.827) (97.71,2.478) (98.47,2.900) (98.89,2.933) (99.14,3.077) (99.30,3.305) (99.40,3.276) (99.49,3.561) (99.56,3.775)};

        \addplot[rred, mark=triangle*] coordinates
          {(50.89,2.008) (64.99,2.395) (72.16,2.760) (76.35,3.139) (78.93,3.646) (80.88,4.131) (82.51,4.495) (83.75,4.900) (84.85,5.507) (85.72,5.807)};
    \end{semilogyaxis}
\end{tikzpicture}
\end{minipage}\hfill
\begin{minipage}[c]{0.39\columnwidth}
    \caption{Latency vs.\ Recall for 10M scale datasets (Web Ads, Yandex T2I~\citep{bigann-benchmark-report21, YandexT2I}, Turing T2I~\citep{T-Bletchley}) show the gap between query efficiency for ID and OOD queries.}
    \label{fig:combined-in-memory}
\end{minipage}
\vspace{-5pt}
\end{figure*}
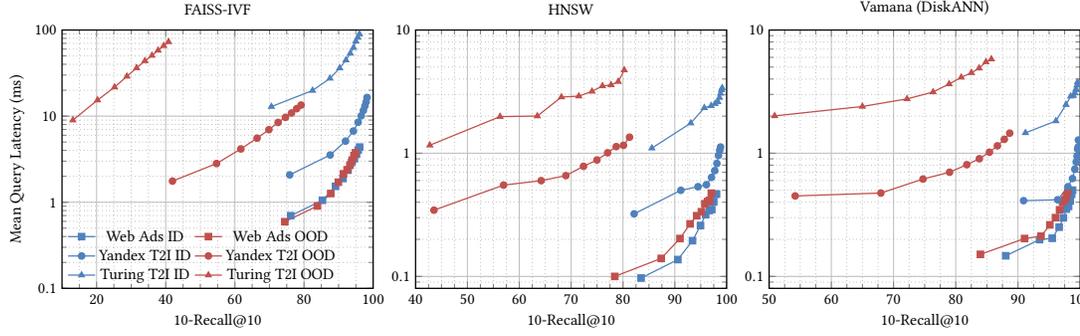

Therefore, it is not surprising that when queries are drawn from a different distribution -- such as in the cross-modal scenario -- indices that optimize for base data distribution exhibit a large drop in accuracy and retrieval efficiency for these ``out-of-distribution'' (OOD) queries. \textbf{\autoref{fig:combined-in-memory}} compares the recall-vs-latency curves of SoTA~\cite{ann-benchmarks} data-dependent algorithms -- \hnsw~\citep{HNSW}, \faiss~\citep{FAISS, IVF-OADCGP} and \previoustool~\citep{DiskANN}-- on three datasets. Two of the data sets are text-to-image while the other corresponds to web advertisements and short query embeddings. There is an order of magnitude gap in the latency required to compute NNs of in-distribution or ID queries (these queries are sampled from the distribution generating the base data set, without replacement) and the OOD queries in both the text-to-image datasets. Such a large drop in latency can make it infeasible to serve accurate results within strict latency budgets needed in web scenarios. To serve multi-modal embeddings and other OOD query sets, there is a need for robust ANNS indices which can adapt to queries drawn from a distribution other than the base data distribution. Hence, this paper asks:
\textbf{Given a small sample set drawn from the query distribution apriori, is it possible to use such a set to build a better ANNS index that works well for OOD queries?}

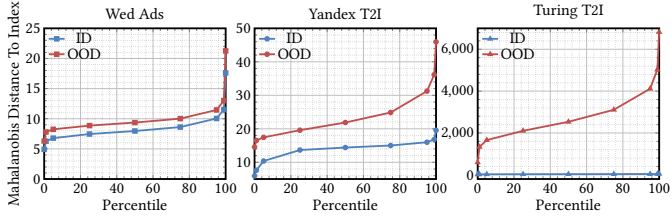
\begin{figure}[t]
\flushleft
\begin{minipage}{\columnwidth}
    \begin{tikzpicture}[scale=0.35,font=\Huge]
    \begin{axis}[ 
        title=Wed Ads,
        width=\linewidth,
        line width=2,
        grid=both,
        grid style={line width=.1pt, draw=gray!10},
        major grid style={line width=.2pt,draw=gray!50},
        minor grid style={line width=.3pt,draw=gray!100, dotted},
        minor tick num=4,
        legend style={nodes={scale=1.0, transform shape}},
        grid style={white},
        xlabel={Percentile},
        ylabel={Mahalanobis Distance To Index},
        y tick label style={
            /pgf/number format/.cd,
            fixed,
            fixed zerofill,
            precision=0
        },
        legend style={at={(0,1)},
        anchor=north west, draw=none, fill=none},
        legend columns=1,
        ymin=0,
        ymax=25,
        xmin=0,
        xmax=100,
    ]
        \addplot[bblue, mark=square*] coordinates
          {(0,4.9256) (1,6.2625) (5,6.7860) (25,7.4640) (50,7.9660) (75,8.6180) (95,10.0469) (99,11.5187) (100,17.6039)};
          \addlegendentry{ID}

        \addplot[rred, mark=square*] coordinates
          {(0,6.2979) (1,7.8037) (5,8.2463) (25,8.8626) (50,9.3664) (75,10.0198) (95,11.4582) (99,13.0180) (100,21.2559)};
          \addlegendentry{OOD}
    \end{axis}
    \hspace{0.33\textwidth}
    \begin{axis}[ 
        title=Yandex T2I,
        width=\linewidth,
        line width=2,
        grid=both,
        grid style={line width=.1pt, draw=gray!10},
        major grid style={line width=.2pt,draw=gray!50},
        minor grid style={line width=.3pt,draw=gray!100, dotted},
        minor tick num=4,
        legend style={nodes={scale=1.0, transform shape}},
        grid style={white},
        xlabel={Percentile},
        y tick label style={
            /pgf/number format/.cd,
            fixed,
            fixed zerofill,
            precision=0
        },
        legend style={at={(0,1)},
        anchor=north west, draw=none, fill=none},
        legend columns=1,
        ymin=5,
        ymax=50,
        xmin=0,
        xmax=100,
    ]
        \addplot[bblue, mark=otimes*] coordinates
          {(0,5.9442) (1,7.6340) (5,10.3789) (25,13.6544) (50,14.4034) (75,15.0087) (95,15.9805) (99,16.8330) (100,19.5370)};
          \addlegendentry{ID}

        \addplot[rred, mark=otimes*] coordinates
          {(0,14.5726) (1,16.5184) (5,17.4400) (25,19.5773) (50,21.8724) (75,24.8648) (95,31.2490) (99,36.1666) (100,45.9716)};
          \addlegendentry{OOD}
    \end{axis}
    \hspace{0.35\textwidth}
    \begin{axis}[ 
        title=Turing T2I,
        width=\linewidth,
        line width=2,
        grid=both,
        grid style={line width=.1pt, draw=gray!10},
        major grid style={line width=.2pt,draw=gray!50},
        minor grid style={line width=.3pt,draw=gray!100, dotted},
        minor tick num=4,
        legend style={nodes={scale=1.0, transform shape}},
        grid style={white},
        xlabel={Percentile},
        y tick label style={
            /pgf/number format/.cd,
            fixed,
            fixed zerofill,
            precision=0
        },
        legend style={at={(0,1)},
        anchor=north west, draw=none, fill=none},
        legend columns=1,
        ymin=-200,
        ymax=7000,
        xmin=0,
        xmax=100,
    ]
        \addplot[bblue, mark=triangle*] coordinates
          {(0,21.4321) (1,24.4644) (5,26.5846) (25,29.9490) (50,32.3278) (75,34.9820) (95,38.6600) (99,41.6059) (100,48.8102)};
          \addlegendentry{ID}

        \addplot[rred, mark=triangle*] coordinates
          {(0,597.0072) (1,1337.9374) (5,1660.1802) (25,2104.2103) (50,2533.8785) (75,3108.8196) (95,4113.4671) (99,5027.1384) (100,6821.7862)};
          \addlegendentry{OOD}
    \end{axis}
\end{tikzpicture}
\end{minipage}
\vspace{-5pt}
\caption{Histogram of Mahalanobis distances for base set-base set (ID) and query set-base set (OOD) for three datasets with minimal, weak and strong OOD properties (left to right).}
\label{fig:combined-mahalanobis-distance}
\end{figure}

\subsection{What Qualifies A Query As OOD?}
\label{sec:ood-preliminaries}
As it is difficult to provide an all-encompassing definition of OOD for all retrieval algorithms, we adopt a simple and natural proxy that seems to correlate highly with the gap in query efficiency between ID and OOD queries for various data-dependent graph and clustering-based ANNS algorithms.
\emph{We say that a set of queries are OOD with respect to the points in the base set if the histogram of Mahalanobis distances between queries and base points is significantly different from the histogram for base point-to-base point distances}. This formulation allows one to categorise query sets as weakly or strongly OOD, through the significance of differences between the histograms. Notice that the gap between ID and OOD histograms in \textbf{\autoref{fig:combined-mahalanobis-distance}} directly correlates with the gap between ID-OOD query efficiency for all three indices in \textbf{\autoref{fig:combined-in-memory}}.\footnote{A lot of work has previously been undertaken for OOD detection using Mahalanobis distance, among other  approaches, especially in the machine learning~\citep{MLOOD-p1,MLOOD-p2,MLOOD-p3}, computer vision~\citep{VisionOOD-p1,VisionOOD-p2} and natural language processing~\citep{NLPOOD-p1} space. The terminology developed for the purpose of OOD categorisation conforms heavily to to these areas, whereas no concrete definitions exist for cross-modal information retrieval to the best of our knowledge.} While this definition is coarse-grained, our empirical measurements on graph algorithms justifies its use, since the edges in graph-based ANN indices tend to ``locally'' approximate the base point set (see \textbf{\autoref{sec:graph-construction}} and \textbf{\autoref{sec:anns-preliminaries}}). Another criteria that correlates with gap between ID and OOD query performance is the cluster radii of the top-10 NNs for a given query (see \textbf{\autoref{fig:combined-cluster-radii}}), which is the minimum radius of a ball enclosing the $10^{\text{th}}$ closest NN.

\subsection {Why Is It Hard To Serve OOD Queries?}
\label{sec:hardness-of-ood}
\begin{figure}[t]
\vspace{-5pt}
\flushleft
\begin{minipage}{\columnwidth}
    \begin{tikzpicture}[scale=0.34,font=\Huge]
    \begin{axis}[ 
        title=Web Ads,
        width=\linewidth,
        line width=2,
        grid=both,
        grid style={line width=.1pt, draw=gray!10},
        major grid style={line width=.2pt,draw=gray!50},
        minor grid style={line width=.3pt,draw=gray!100, dotted},
        minor tick num=4,
        legend style={nodes={scale=1.0, transform shape}},
        grid style={white},
        xlabel={Percentile},
        ylabel={10-NN Cluster Radius},
        y tick label style={
            /pgf/number format/.cd,
            fixed,
            fixed zerofill,
            precision=0
        },
        legend style={at={(0,1)},
        anchor=north west, draw=none, fill=none},
        legend columns=1,
        ymin=0,
        ymax=150,
        xmin=0,
        xmax=100,
    ]
        \addplot[bblue, mark=square*] coordinates
          {(0,1.6941) (1,13.6987) (5,25.3141) (25,43.2284) (50,55.7854) (75,69.7301) (95,90.2928) (99,100.4394) (100,130.0189)};
          \addlegendentry{ID}

        \addplot[rred, mark=square*] coordinates
          {(0,7.2166) (1,31.8482) (5,38.9820) (25,52.5880) (50,64.6539) (75,78.8916) (95,97.5175) (99,106.4784) (100,141.1305)};
          \addlegendentry{OOD}
    \end{axis}
    \hspace{.33\textwidth}
    \begin{axis}[ 
        title=Yandex T2I,
        width=\linewidth,
        line width=2,
        grid=both,
        grid style={line width=.1pt, draw=gray!10},
        major grid style={line width=.2pt,draw=gray!50},
        minor grid style={line width=.3pt,draw=gray!100, dotted},
        minor tick num=4,
        legend style={nodes={scale=1.0, transform shape}},
        grid style={white},
        xlabel={Percentile},
        y tick label style={
            /pgf/number format/.cd,
            fixed,
            fixed zerofill,
            precision=1
        },
        legend style={at={(0,1), font=\normalsize},
        anchor=north west, draw=none, fill=none},
        legend columns=1,
        ymin=0,
        ymax=1.2,
        xmin=0,
        xmax=100,
    ]
        \addplot[bblue, mark=otimes*] coordinates
          {(0,0.0844) (1,0.2048) (5,0.2684) (25,0.4029) (50,0.4985) (75,0.5854) (95,0.6858) (99,0.7456) (100,0.9647)};
          \addlegendentry{ID}

        \addplot[rred, mark=otimes*] coordinates
          {(0,0.1332) (1,0.3280) (5,0.4391) (25,0.5771) (50,0.6623) (75,0.7382) (95,0.8437) (99,0.9250) (100,1.1476)};
          \addlegendentry{OOD}
    \end{axis}
    \hspace{.33\textwidth}
    \begin{axis}[ 
        title=Turing T2I,
        width=\linewidth,
        line width=2,
        grid=both,
        grid style={line width=.1pt, draw=gray!10},
        major grid style={line width=.2pt,draw=gray!50},
        minor grid style={line width=.3pt,draw=gray!100, dotted},
        minor tick num=4,
        legend style={nodes={scale=1.0, transform shape}},
        grid style={white},
        xlabel={Percentile},
        y tick label style={
            /pgf/number format/.cd,
            fixed,
            fixed zerofill,
            precision=1
        },
        legend style={at={(0,1)},
        anchor=north west, draw=none, fill=none},
        legend columns=1,
        ymin=0,
        ymax=1.2,
        xmin=0,
        xmax=100,
    ]
        \addplot[bblue, mark=triangle*] coordinates
          {(0,0.0689) (1,0.3986) (5,0.4811) (25,0.5849) (50,0.6474) (75,0.7043) (95,0.7769) (99,0.8285) (100,0.9937)};
          \addlegendentry{ID}

        \addplot[rred, mark=triangle*] coordinates
          {(0,0.2554) (1,0.5985) (5,0.6715) (25,0.7577) (50,0.8093) (75,0.8569) (95,0.9275) (99,0.9801) (100,1.1601)};
          \addlegendentry{OOD}
    \end{axis}
\end{tikzpicture}
\end{minipage}
\vspace{-5pt}
\caption{Cluster radii of top-10 NNs of index point (ID) and query (OOD) sample sets. }
\label{fig:combined-cluster-radii}
\end{figure}
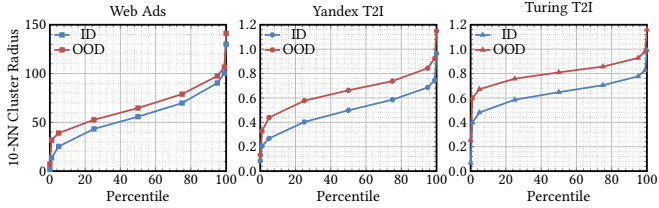

Compared to the ID queries, OOD queries pose two major problems: 

\textbf{Poor ``Clusterability'' of the $k$-NNs of a Query:} The large gap in latency for OOD queries can be explained by the fact that the search algorithm needs to look at a lot more points in the index to hit the same recall target as ID queries.
This discrepancy can be explained by looking at the ``clusterability'' of the $k$-NN nodes pertaining to an ID query and an OOD query. As evident from \textbf{\autoref{fig:combined-cluster-radii}}, OOD queries have their corresponding $k$-NNs spread over a larger volumetric space than their ID counterparts. This has different bearings for different types of ANNS algorithms.

For clustering-based solutions, such as \faiss, this means that the $k$-NNs of an OOD query do not lie entirely in the clusters deemed closest to the query. Thus, a larger number of clusters need to be probed for finding all of the $k$-NNs, leading to a higher latency.

\begin{figure}
    \vspace{-5pt}
    \centering
    \includegraphics[scale=0.15]{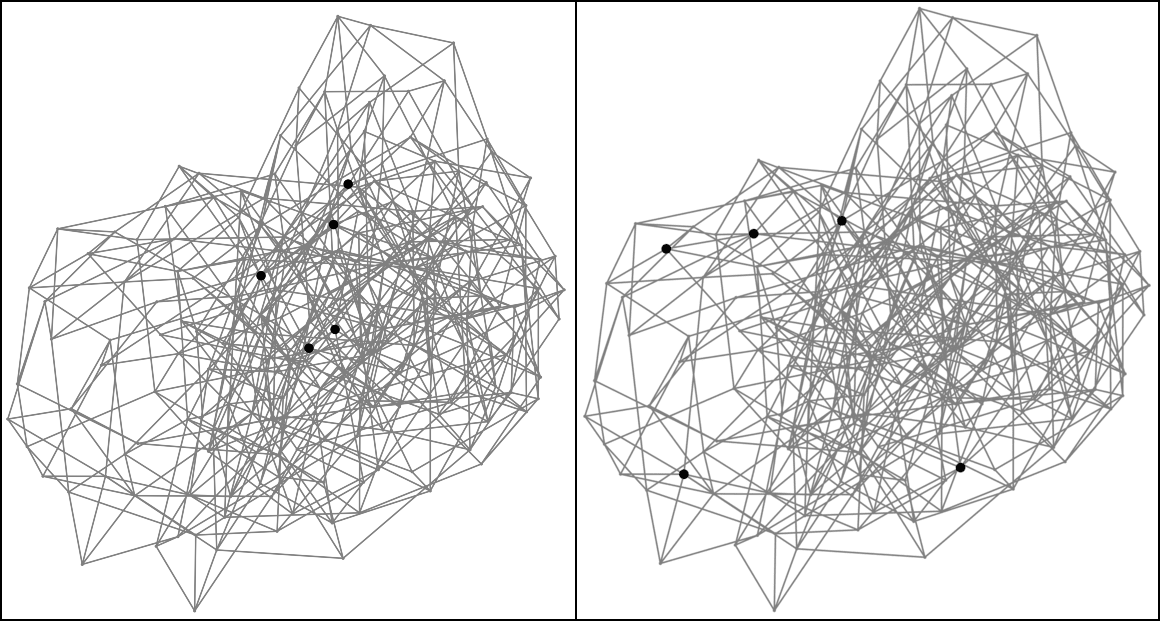}
    \vspace{-5pt}
    \caption{(Left) Top-5 NNs (in black) of an ID query over a random subgraph of the Yandex Text-to-Image-1B dataset. (Right) Top-5 NNs of an OOD query over the same subgraph. Interconnected vertices appear closer to each other.}
    \label{fig:IDOODExample}
\end{figure}

For graph-structured ANNS indices, this translates to poor connectivity among the $k$-NNs of a given query -- \textbf{\autoref{fig:IDOODExample}} demonstrates this phenomenon on a $200$ point subgraph of the Yandex T2I dataset, where the $k$-NNs of an ID query are locally well connected whereas that of an OOD query are much less connected for the graph constructed by the \vamana~\citep{DiskANN} indexing algorithm. 
Poor edge connectivity among the $k$-NNs of a query directly translates into poor recall scores, since the greedy search is prone to running into a ``local minima'' by visiting only a fraction of the true $k$-NNs.

\textbf{Misaligned Loss of Quantization Schemes:} Quantization schemes such as Product Quantization (PQ~\citep{PQ}) and Optimized Product Quantization (OPQ~\citep{OPQ}) -- that are critical to large billion-scale indices~\citep{bigann-benchmark-report21} like \faisspq~\citep{FAISS, IVF-OADCGP} and \previoustool~\citep{DiskANN} -- are extremely inaccurate for OOD queries. These quantization methods allow the indexing of billion+ high-dimensional vectors that do not fit in DRAM natively ($100$M points in $1024$ dimensions needs $384$GBs of memory) on commodity machines with little DRAM (e.g. $64$GB). They also allow for faster, albeit approximate, distance comparisons owing to their smaller CPU and memory bandwidth requirements leading to algorithms with low latency. However, classical  formulations of these schemes \emph{attempt to minimize $\ell_2$ distance distortions between points in the base data distribution and their quantized vectors}. When used to compute distance estimates between queries drawn from a different distribution and the base dataset, this leads to extreme distortions in relative ranking of distances from the query to its nearest point -- \emph{in fact, the actual nearest neighbor may not be in the top 10 candidates in terms of quantized distances to the query!} Therefore, even if we attempt to correct this by re-ranking the algorithm's limited candidate pool with higher precision vectors, the recall may not improve since the top neighbors may not even show in the candidate pool.





\subsection{Our Contributions}
This paper develops the following techniques towards addressing these challenges: 
\begin{enumerate}
    \item We present \robustvamana, which improves upon the \vamana graph construction algorithm of \previoustool, and provides superior search latencies for OOD queries (\textbf{\autoref{sec:graph-construction}}).
    \item We argue, theoretically and empirically, how the existing PQ compression schemes can cause large distortions in distance estimates for OOD queries. We therefore introduce a new formulation, called Accurate PQ (or APQ), that improves performance for OOD queries (\textbf{\autoref{sec:accurate-product-quantization}}).
    \item We present \parallelgorder, a parallel and scalable graph reordering algorithm based on a previous work~\citep{Gorder}, which reduces I/O requests to SSDs, the primary performance bottleneck for large-scale external memory indices. This provides improved latency for SSD-based indices (\textbf{\autoref{sec:graph-reordering}}).
    \item We put them all together to build billion-scale indices, and demonstrate efficiency gains for OOD queries by upto $40\%$ in terms of mean latency or, alternatively, upto $15\%$ in terms of recall over prior SoTA~\citep{bigann-benchmark-report21}.
\end{enumerate}


\subsection{Notation}
We denote the dataset of base points as $\mathcal{X}$, where $i^{\text{th}}$ point has coordinates $x_{i} \in \mathbb{R}^{D}$. We consider directed graphs with vertices corresponding to points in $\mathcal{X}$, and edges $\mathcal{E}$ between them. We refer to such graphs as $\mathcal{G} = (\mathcal{X}, \mathcal{E})$ with slight notational overhead. We denote the sample set of query points as $\mathcal{Q}$, where the $i^{\text{th}}$ query point has coordinates $q_{i} \in \mathbb{R}^{D}$.


\subsection{Related Work}

PQ and OPQ have been the seminal works in the area of vector compression for fast distance estimates and compact memory footprint. They have also inspired algorithms such as LOPQ~\citep{LOPQ}, LSQ~\citep{LSQ-v1}, LSQ++~\citep{LSQ-v2}, ERVQ~\citep{ERVQ}, CQ~\citep{CQ} and AQ~\citep{AQ}, and learnt decoder-based techniques such as UNQ~\citep{UNQ} and Learnt AQ~\citep{mathijs-decoder}, all of which have pushed the envelope on the accuracy of the distance estimates.

\section{\robustvamana Graph Construction}
\label{sec:graph-construction}
In this section, we focus on graph-based ANNS indices and present techniques which substantially improve over existing methods for the OOD setting. We focus on the \vamana graph construction heuristic, which is an integral part of the DRAM/SSD hybrid \previoustool system, as our goal is to improve the SoTA for very large scale ANNS scenarios, thereby necessitating SSD-resident indices. An interested reader may refer to \textbf{\autoref{sec:anns-preliminaries}} for preliminaries.

We first explain the drawbacks existing algorithms face with OOD queries by briefly describing how such algorithms build and search graph indices. Many popular graph-based indices such as \vamana, \hnsw, \nsg~\citep{NSG}, etc. build so-called \emph{navigable graphs} which employ a simple \emph{greedy heuristic} at search time -- the search starts with a \emph{candidate list} $\mathcal{L}$ initialized with a starting vertex $s$. At any intermediate step, we choose a candidate $p \in \mathcal{L}$, and add its graph out-neighbors $N_{\text{out}}(p)$ to $\mathcal{L}$. To prevent search complexity from exploding, we truncate the list $\mathcal{L}$ to a size of $L$ by retaining the $L$ closest candidates to the query. The search stops once the list does not change, i.e., we arrive at a locally optimal set of candidates.

At a high level, the graphs are then typically constructed in a manner so that every base point can be reached starting from $s$ using the greedy search heuristic. Of course there are nuances in the manner in which the algorithms ensure the degree of the graph is bounded, and the specifics of how to add links to ensure good search performance. However, to our knowledge, all algorithms construct the graph index by only considering the base dataset, completely oblivious of the query distribution. Indeed, this can pose major challenges in the search navigability for OOD queries, as highlighted in \textbf{\autoref{sec:hardness-of-ood}}.



We address this problem by giving access to a small sample, say $\sim1\text{-}2\%$ of the base dataset size, of OOD queries to the index construction algorithm. Like \vamana, our \robustvamana algorithm incrementally and iteratively builds the navigable graph, with the $i^{\text{th}}$ iteration ensuring that the graph can connect point $x_{i}$. A crucial difference is that we include the query sample points in this process. However, a query point will only explore candidates from the base dataset and add links to the closest $R$ base points, whereas a base point will add links to other base points as well as the query points. 

Then using these base / query edges we have identified in the above process, we add cross-links between base points which co-occur as the out-neighbors for different query points, subject to the degree threshold of the graph. This has the intended effect of inter-connecting all of the NNs of a sample query. Keeping the query points embedded in the index is a no go, since they can saturate the candidate list during search, by virtue of being closer to the evaluation queries. As such, they can kick the true NNs from the base set out of the candidate list, leading to poor recall. Hence, once the cross-links are added using \robuststitch (\textbf{\autoref{alg:robust-stitch}}), we remove these sample query points from the index. The complete specification of \robustvamana is provided in \textbf{\autoref{alg:robust-vamana}}.

\subsection{Improvement Over \vamana}
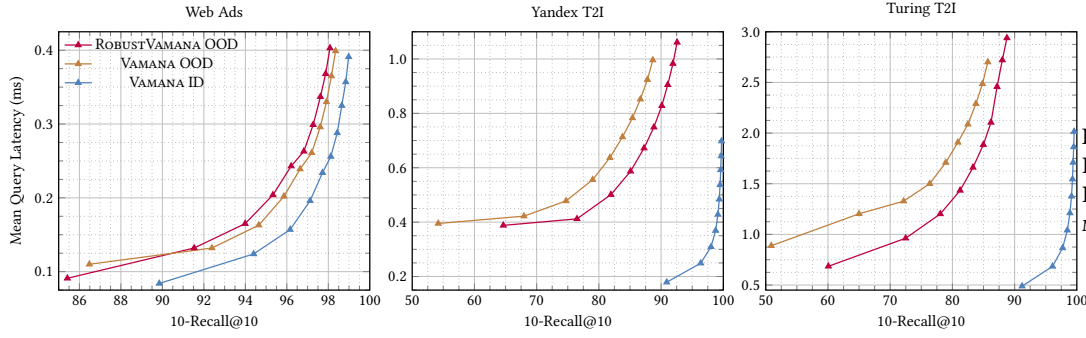
\begin{figure*}
\flushleft
\begin{minipage}[c]{\columnwidth}
    \begin{tikzpicture}[scale=0.6,font=\large]
    \begin{axis}[
        title=Web Ads,
        width=\linewidth,
        line width=0.8,
        grid=both,
        grid style={line width=.1pt, draw=gray!10},
        major grid style={line width=.2pt,draw=gray!50},
        minor grid style={line width=.3pt,draw=gray!100, dotted},
        minor tick num=3,
        tick label style={font=\large},
        legend style={nodes={scale=1.0, transform shape}},
        label style={font=\large},
        legend image post style={mark=triangle*},
        grid style={white},
        xlabel={10-Recall@10},
        ylabel={Mean Query Latency (ms)},
        y tick label style={
            /pgf/number format/.cd,
            fixed,
            fixed zerofill,
            precision=1
        },
        legend style={at={(0,1), font=\normalsize}, anchor=north west, draw=none, fill=none},
        legend columns=1,
        ymin=0.075,
        ymax=0.425,
        xmin=85,
        xmax=100,
    ]
        \addplot[purple, mark=triangle*] coordinates
          {(85.42,0.091) (91.53,0.132) (93.99,0.165) (95.33,0.204) (96.21,0.243) (96.82,0.263) (97.27,0.299) (97.62,0.337) (97.87,0.368) (98.08,0.403)};
          \addlegendentry{\robustvamana OOD}

        \addplot[brown, mark=triangle*] coordinates
          {(86.48,0.110) (92.39,0.132) (94.65,0.163) (95.86,0.202) (96.64,0.239) (97.20,0.261) (97.61,0.296) (97.91,0.330) (98.15,0.365) (98.35,0.399)};
          \addlegendentry{\vamana OOD}
        
        \addplot[bblue, mark=triangle*] coordinates
          {(89.85,0.084) (94.40,0.124) (96.16,0.157) (97.12,0.196) (97.72,0.234) (98.12,0.256) (98.43,0.288) (98.65,0.325) (98.84,0.357) (98.98,0.391)};
          \addlegendentry{\vamana ID}
    \end{axis}
    \hspace{4.7cm}
    \begin{axis}[
        title=Yandex T2I,
        width=\linewidth,
        line width=0.8,
        grid=both,
        grid style={line width=.1pt, draw=gray!10},
        major grid style={line width=.2pt,draw=gray!50},
        minor grid style={line width=.3pt,draw=gray!100, dotted},
        minor tick num=3,
        tick label style={font=\large},
        legend style={nodes={scale=1.0, transform shape}},
        label style={font=\large},
        legend image post style={mark=triangle*},
        grid style={white},
        xlabel={10-Recall@10},
        y tick label style={
            /pgf/number format/.cd,
            fixed,
            fixed zerofill,
            precision=1
        },
        legend style={at={(0,1), font=\normalsize}, anchor=north west, draw=none, fill=none},
        legend columns=1,
        ymin=0.15,
        ymax=1.10,
        xmin=50,
        xmax=100,
    ]
        \addplot[purple, mark=triangle*] coordinates
          {(64.64,0.388) (76.48,0.412) (81.93,0.501) (85.09,0.587) (87.26,0.672) (88.84,0.749) (90.11,0.828) (91.07,0.905) (91.90,0.983) (92.57,1.061)};

        \addplot[brown, mark=triangle*] coordinates
          {(54.19,0.395) (67.96,0.422) (74.73,0.478) (79.00,0.556) (81.77,0.637) (83.79,0.713) (85.40,0.783) (86.67,0.852) (87.80,0.924) (88.67,0.996)};
        
        \addplot[bblue, mark=triangle*] coordinates
          {(90.90,0.179) (96.37,0.249) (98.02,0.309) (98.76,0.369) (99.13,0.428) (99.35,0.484) (99.48,0.537) (99.58,0.592) (99.66,0.643) (99.71,0.698)};
    \end{axis}
    \hspace{4.7cm}
    \begin{axis}[
        title=Turing T2I,
        width=\linewidth,
        line width=0.8,
        grid=both,
        grid style={line width=.1pt, draw=gray!10},
        major grid style={line width=.2pt,draw=gray!50},
        minor grid style={line width=.3pt,draw=gray!100, dotted},
        minor tick num=3,
        tick label style={font=\large},
        legend style={nodes={scale=1.0, transform shape}},
        label style={font=\large},
        legend image post style={mark=triangle*},
        grid style={white},
        xlabel={10-Recall@10},
        y tick label style={
            /pgf/number format/.cd,
            fixed,
            fixed zerofill,
            precision=1
        },
        legend style={at={(0,1), font=\normalsize}, anchor=north west, draw=none, fill=none},
        legend columns=1,
        ymin=0.45,
        ymax=3.0,
        xmin=50,
        xmax=100,
    ]
        \addplot[purple, mark=triangle*] coordinates
          {(60.07,0.685) (72.48,0.962) (78.05,1.202) (81.24,1.434) (83.31,1.661) (84.97,1.885) (86.20,2.105) (87.18,2.458) (88.02,2.721) (88.76,2.939)};

        \addplot[brown, mark=triangle*] coordinates
          {(50.89,0.888) (64.99,1.203) (72.16,1.328) (76.35,1.500) (78.93,1.709) (80.88,1.908) (82.49,2.087) (83.78,2.290) (84.83,2.486) (85.69,2.701)};
        
        \addplot[bblue, mark=triangle*] coordinates
          {(91.20,0.489) (96.09,0.684) (97.71,0.866) (98.47,1.041) (98.89,1.211) (99.14,1.376) (99.30,1.545) (99.40,1.710) (99.49,1.863) (99.56,2.015)};
    \end{axis}
\end{tikzpicture}
\end{minipage}\hfill
\begin{minipage}[c]{0.4\columnwidth}
    \caption{(Left to right) Latency vs.\ Recall for \robustvamana and \vamana at 10M scale.}
    \label{fig:combined-graph}
\end{minipage}
\vspace{-5pt}
\end{figure*}

\textbf{\autoref{fig:combined-graph}} summarises our results, where we observe an improvement of $4\text{-}10$ recall points, for the same latency target across the text-to-image datasets. For the Web Ads dataset, we observe a negligible change in latency vs.\ recall. This suggests that \robustvamana improves OOD query search performance, without interfering with the ID query search performance. Since the query sample size is merely $1\text{-}2\%$ of the base dataset, the overall build-phase time cost and peak RAM footprints increase by the same amount, while the search-phase peak RAM and disk footprints remain unchanged.



\begin{algorithm}
\caption{\greedysearch($s$, $x_{q}$, $\tau$, $\mathcal{L}$)}
\label{alg:greedy-search}
\DontPrintSemicolon
\small
\KwData{Start node $s$, query vector $x_{q}$, boolean $\tau$, search list size $\mathcal{L}$.}
\KwResult{Result set $\mathcal{L}$ containing near neighbors, and a set $\mathcal{V}$ containing all visited nodes.}
\Begin{
    \nonl $\text{Initialize sets } \mathcal{L} \leftarrow \{s\} \text{ and } \mathcal{V} \leftarrow \emptyset.$\;
    \While{$\mathcal{L} \setminus \mathcal{V} \neq \emptyset$}{
        \nonl $\text{Let } p^{*} \leftarrow \text{arg min}_{p \in \mathcal{L} \setminus \mathcal{V}} \|x_{p} - x_{q}\|$\;
        \nonl $\mathcal{V} \leftarrow \mathcal{V} \cup \{p^{*}\}$\;
        \eIf{$\tau$}{
            \nonl $\mathcal{L} \leftarrow \mathcal{L} \cup \{v\text{ }|\text{ }v \in N_{\text{out}}(p^{*}) \text{, } v \in \mathcal{X}\}$\;
        }{
            \nonl $\mathcal{L} \leftarrow \mathcal{L} \cup N_{\text{out}}(p^{*})$\;
        }
        \If{$|\mathcal{L}| > L$}{
            \nonl $\text{Update } \mathcal{L} \text{ with closest } L \text{ nodes to } x_{q}.$\;
        }
    }
    \KwRet $[\mathcal{L}; \mathcal{V}]$\;
}
\end{algorithm}
\begin{algorithm}
\caption{\robustprune($p$, $\mathcal{V}$, $\alpha$, $R$)}
\label{alg:robust-prune}
\DontPrintSemicolon
\small
\KwData{Point $p \in \mathcal{X}$, candidate set $\mathcal{V}$, parameter $\alpha \geq 1$, outdegree $R$.}
\KwResult{$\mathcal{G}$ is modified by setting at most $R$ out-neighbors for $p$.}
\Begin{
    \nonl $\mathcal{V} \leftarrow \mathcal{V} \cup N_{\text{out}}(p) \setminus \{p\}$\;
    \nonl $N_{\text{out}}(p) \leftarrow \emptyset$\;
    \While{$\mathcal{V} \neq \emptyset$}{
        \nonl $p^{*} \leftarrow \text{arg min}_{p^{\prime} \in \mathcal{V}} \|x_{p} - x_{p^{\prime}}\|$\;
        \nonl $N_{\text{out}}(p) \leftarrow N_{\text{out}}(p) \cup \{p^{*}\}$\;
        \If{$|N_{\text{out}}(p)| = R$}{
            \nonl $\textbf{break}$\;
        }
        \For{$p^{\prime} \in \mathcal{V}$}{
            \If{$\alpha \cdot \|x_{p^{*}} - x_{p^{\prime}}\| \leq \|x_{p} - x_{p^{\prime}}\|$}{
            \nonl $\text{Remove } p^{\prime} \text{ from } \mathcal{V}.$\;
            }
        }
    }
}
\end{algorithm}


\begin{algorithm}
\caption{\robuststitch($p$, $\mathcal{W}$, $\mathcal{S}$)}
\label{alg:robust-stitch}
\DontPrintSemicolon
\small
\KwData{Point $p \in \mathcal{Q}$, its in-neighbors $\mathcal{W}$ and spare space counts $\mathcal{S}$.}
\KwResult{$\mathcal{G}$ is modified by interconnecting out-neighbors of $p$.}
\Begin{
    \For{$v \in \mathcal{W}$}{
        \nonl $\text{Remove } p \text{ from } N_{\text{out}}(v).$\;
        \nonl $N_{\text{out}}(v) \leftarrow N_{\text{out}}(v) \cup \{\text{closest } \mathcal{S}[v] \text{ elements from } N_{\text{out}}(p) \setminus \{v\}\} $\;
    }
}
\end{algorithm}
\begin{algorithm}
\caption{\robustvamana Indexing Algorithm}
\label{alg:robust-vamana}
\DontPrintSemicolon
\small
\SetKwFunction{FGreedySearch}{GreedySearch}
\SetKwFunction{FRobustPrune}{RobustPrune}
\SetKwFunction{FRobustStitch}{RobustStitch}
\KwData{Parameters $\alpha_{1}$, $\alpha_{2}$, $L$ and $R$.}
\KwResult{Directed graph $\mathcal{G}$ over $\mathcal{X}$ with out-degree $\leq R$.}
\Begin{
    \nonl $\text{Let } s \text{ denote the point closest to the centroid of the dataset } \mathcal{X}\text{.}$\;
    \For{$\alpha \in \{\alpha_{1}, \alpha_{2}\}$}{
        \For{$i \in (\mathcal{X}, \mathcal{Q})$}{
            \eIf{$i \in \mathcal{X}$}{
                \nonl $\text{Let } [\mathcal{L}; \mathcal{V}] \leftarrow$ \FGreedySearch{$s, x_{i}, \text{false}, L$}\;
                \nonl $\text{Update } N_{\text{out}}(i) \text{ using }$ \FRobustPrune{$i, \mathcal{V}, \alpha, R$}.\;
            }{
                \nonl $\text{Let } [\mathcal{L}; \mathcal{V}] \leftarrow$ \FGreedySearch{$s, q_{i}, \text{true}, L$}\;
                \nonl $\text{Update } N_{\text{out}}(i) \text{ using closest } R \text{ nodes from } \mathcal{L}.$\;
            }
            \For{$j \in N_{\text{out}}(i) \wedge j \in \mathcal{X}$}{
                \nonl $\text{Update } N_{\text{out}}(j) \leftarrow N_{\text{out}}(j) \cup i$\;
                \If{$|N_{\text{out}}(j)| > R$}{
                    \nonl $\text{Update } N_{\text{out}}(j) \text{ using }$ \FRobustPrune{$j, N_{\text{out}}(j), \alpha, R$}.\;
                }
            }
        }
    }
    \nonl $\text{Let } \mathcal{S} \leftarrow \{s_{i}\text{ }|\text{ }s_{i} \leftarrow (R - |N_{\text{out}}(i)|),\text{ }i \in \mathcal{X}\}$\;
    \For{$p \in \mathcal{X}$}{
        \nonl $k \leftarrow \{v\text{ }|\text{ }v \in N_{\text{out}}(p)\text{, }\text{ }v \in \mathcal{Q}\}$\;
        \nonl $\mathcal{S}[p] \leftarrow \lfloor\frac{\mathcal{S}[p]}{|k|}\rfloor + 1$\;
    }
    \For{$p \in \mathcal{Q}$}{
        \nonl $\text{Let } \mathcal{W} \leftarrow \{v\text{ }|\text{ }p \in N_{\text{out}}(v)\text{, }\text{ }v \in \mathcal{X}\}$\;
        \nonl $\text{Run }$ \FRobustStitch{$p, \mathcal{W}, \mathcal{S}$} $\text{to update the graph.}$\;
    }
}
\end{algorithm}

\section{Query-Aware Product Quantization}
\label{sec:accurate-product-quantization}
We now turn our attention to devising improved \emph{quantization schemes} by making them query aware. Product Quantization (PQ) and its variants like OPQ~\citep{OPQ} are a very popular class of quantization methods that enable large-scale ANNS on a modest memory footprint.
At a high level, these methods compress $D$-dimensional vector data into a certain number of bytes per vector as follows: the $D$ dimensions are divided into $M$ \emph{chunks} of $D/M$ dimensions each. Within each chunk in $D/M$-dimensional subspace, an algorithm \emph{learns} $K$ (usually taken to be $K=256$) representative pivots/centers that best \emph{approximate} the original coordinates in the chunk across the dataset. Typically, this step employs the $K$-means heuristic over the base vectors restricted to that subspace. Then, each vector can be represented by the closest pivot/center within the chunk, thereby requiring only $\log_2 K$ bits to encode, which is $1$ byte for $K = 256$. Following this process independently for each chunk, we can approximate any vector using a cross product of pivots in the chunk-wise sub-spaces using a total of $M$ bytes. Schemes like OPQ, LOPQ and others build on top of this basic primitive.

Our main contribution in this section is to essentially redesign the quantization process within each chunk by making it aware of the target query distribution. An interested reader may further refer to \textbf{\autoref{sec:pq-preliminaries}}, \textbf{\autoref{sec:pq-pitfalls}} and \textbf{\autoref{sec:apq-ip-analysis}}.


\subsection{Problem with Existing Methods}
Consider an existing quantization scheme and suppose that for some base vector $x_{i}$, it's restriction $x^{j}_{i}$ within a chunk $j$ is assigned to a representative pivot $\mu$. Also let $q^{j}$ denote the restriction of a query $q$ to the this chunk. Then we have:
\begin{equation}
\label{eq:1}
    \|q^{j} - x^{j}_{i}\|^{2}_{2}
    = \|q^j - \mu + \mu - x^{j}_{i}\|^{2}_{2} 
    = \|q^j - \mu\|^{2}_{2} + \|x^j_{i} - \mu\|^{2}_{2}  - 2 \langle q^j - \mu, x^j_{i} - \mu \rangle
\end{equation}


\noindent Thus, chunk-wise error between actual and estimated distance is:
\begin{eqnarray}
\label{eq:2}
    \underbrace{\|q^{j} - x^{j}_{i}\|^{2}_{2}}_{\text{Actual}} - \underbrace{\|q^{j} - \mu\|^{2}_{2}}_{\text{Estimated}} & = \langle x^j_{i} + \mu - 2q^j,  x^j_{i} - \mu \rangle
\end{eqnarray}


A positive value of the above term signifies that the base point appears closer to the query than it actually is, and vice versa. To provide intuition on why existing methods are not very performant on OOD datasets, we re-organize this distortion as:
\begin{eqnarray}
\label{eq:3}
    \langle x^{j}_{i} + \mu - 2q^{j},  x^{j}_{i} - \mu \rangle = 2\|\frac{x^{j}_{i} + \mu}{2} - q^{j}\|_{2} \|x^{j}_{i} - \mu\|_{2}\cos{\theta}
\end{eqnarray}
where $\theta$ is the angle between the above two inner product vectors. The crucial issue is that \emph{existing PQ schemes only aim to minimize the squared euclidean norm of the second multiplicative term $\|x^j_{i} - \mu\|_{2}$}, which is optimized by finding by the $K$ pivots through classical $K$-means method and assigning each point to the nearest pivot. 

In order to remedy this issue, we propose Accurate Product Quantization (APQ) which \emph{chooses to directly optimize for the true distortion} in~\autoref{eq:2}. More formally, for each sample query $q \in \mathcal{Q}$, let $H_T(q)$ denote the set of closest $T$ vectors to $q$ for a parameter $T$. Then, for each base point $x_{i} \in \mathcal{X}$, we define a set of relevant queries as follows: $\mathcal{Q}_{x_{i}} = \{q : x_{i} \in H_{T}(q) \}$, and if $|\mathcal{Q}_{x_{i}}| > \phi$, we refine $\mathcal{Q}_{x_{i}}$ to contain the closest $\phi$ queries. 

For each chunk, we are now ready to define our loss function $\mathcal{L}_{\text{APQ-L2}}$ for assigning the restriction $x^{j}_{i}$, of $x_i$ in the $j^{\text{th}}$ chunk, to a candidate pivot vector $\mu$:
\begin{flalign}
\label{eq:4}
    & \mathcal{L}_{\text{APQ-L$2$}}(x^j_{i}, \mu, \mathcal{Q}_{x_{i}}) \nonumber \\ & = \frac{1}{|\mathcal{Q}_{x_{i}}|} \sum_{q \in \mathcal{Q}_{x_{i}}}
        \begin{cases}
            -\langle x^j_{i} + \mu - 2q^j,  x^j_{i} - \mu \rangle, & \text{if}\ x_{i} \in \text{H}_{T'}(q) \\
            |\langle x^j_{i} + \mu - 2q^j,  x^j_{i} - \mu \rangle|, & \text{otherwise}
        \end{cases} &&
\end{flalign}
where, $\mu \in \mathbb{R}^{\frac{D}{M}}$ is the representative pivot (to be learnt) that $x^{j}_{i}$ will be assigned to in the quantization. This formulation is inspired from prior work employing contrastive losses~\citep{CL-p1, CL-p2, CL-p3, CL-p4}. In the first case, the optimizer pushes both the negative and positive distortions in the positive direction for a select subset of relevant queries which are known to be really close (as $T'<<T$). In the second case, the optimizer minimizes the magnitude of the distortions themselves.

The overall optimization for the given chunk $j$ is to choose $K$ pivots $\{\mu_1, \mu_2, \ldots, \mu_K\}$ and assign each base vector $x^{j}_{i}$ to one of these pivots $\mu_{\sigma(i)}$ to minimize the total loss $\sum_{i} \mathcal{L}_{\text{APQ-L$2$}}(x^j_{i}, \mu_{\sigma(i)}, \mathcal{Q}_{x_{i}})$. We optimize this loss objective in~\autoref{eq:4} through an alternating minimization approach. \textbf{\autoref{alg:chunk-wise-accurate-pq}} and \textbf{\autoref{alg:accurate-pq-codebook-learning}} summarise our approach where we partition our base dataset into $M$ chunks, and learn a set of $K$ independent pivots for each chunk. We initiate the learning process for a chunk by choosing a random set of $K$ pivot points, and follow up with a membership assignment and pivot update steps in an alternating fashion. The learnt set of pivots pertaining to the $j^{\text{th}}$ chunk form a dictionary $C_{j}$, and the final set of $M$ dictionaries form our learnt codebook $\mathcal{C}$. This codebook is then used to encode points in the index.

To illustrate the power and generality of this approach, we also deploy this approach within the OPQ framework which learns a product quantization of the vectors after a suitable rotation of the data and queries. We don't modify the steps which learn the rotation, but employ the APQ algorithm in lieu of the steps which learn the PQ codebook. We refer to the resulting scheme as AOPQ.

\begin{algorithm}[t]
\caption{Chunk-Wise Accurate PQ Algorithm}
\label{alg:chunk-wise-accurate-pq}
\DontPrintSemicolon
\small
\SetKwInput{KwInit}{Initialization Step}
\SetKwInput{KwPvtUpd}{Pivot Update Step}
\KwData{The restriction $\mathcal{X}^{j}$ of the given data $\mathcal{X}$ to a chunk of $D/M$ dimensions, and number of pivots $K$.}
\KwResult{Learnt pivots pertaining to the chunk.}
\KwInit{Sample $K$ random points from $\mathcal{X}^{j}$, and assign them as initial pivots $\mu_{1}, \dots, \mu_{K}$.}
\Begin{
    \nonl \textbf{Membership Assignment Step:} Find the most suitable pivot $\Tilde{\mu}$ for each base point chunk $x^j_{i},$ by looking at all possible choices for $\mu$:\;
    \begin{align*}
        \Tilde{\mu} = \underset{\mu \in \{\mu_{1}, \dots, \mu_{K}\}}{\operatorname{arg min}} \mathcal{L}_{\text{APQ}}(x^{j}_{i}, \mu, \mathcal{Q}_{x_{i}}) &&
    \end{align*}\;
    \nonl \textbf{Pivot Update Step:} For every pivot $\mu_{i}$, let $X_{\mu_{i}}$ be the set of base points assigned to it. Then update $\mu_{i}$ by running gradient descent to optimize the loss:\;
    \begin{align*}
        \mu_{i} = \underset{\mu \in \mathbb{R}^{D/M}}{\operatorname{arg min}} \sum_{x^j_{i} \in X_{\mu_{i}}} \mathcal{L}_{\text{APQ}}(x^{j}_{i}, \mu, \mathcal{Q}_{x_{i}}) &&
    \end{align*}\;
    \nonl Repeat \textbf{Membership Assignment Step} and \textbf{Pivot Update Step} until either convergence or maximum number of iteration is reached.\;
}
\end{algorithm}
\begin{algorithm}[t]
\caption{Accurate PQ Codebook Learning Algorithm}
\label{alg:accurate-pq-codebook-learning}
\DontPrintSemicolon
\small
\SetKwFor{ParallelFor}{parallel for}{do}{endfor}
\KwData{Number of pivots per chunk $K$ and number of chunks $M$.}
\KwResult{Learnt codebook $\mathcal{C} = \{C_{1}, C_{2}, \dots, C_{M}\}$ of pivots pertaining to all chunks.}
\Begin{
    \ParallelFor{$j \in \{0, \dots, M - 1\}$}{
        \nonl \textbf{Initialization Step:} Create a chunk dataset $\mathcal{X}^{j}$ by splitting each data point $x_{i}$ into $M$ chunks, each of dimension $\frac{D}{M}$, such that for a chunk $j$:\;
        \begin{align*}
            x^{j}_{i} = (x_{i, 1 + j * \frac{D}{M}}, x_{i, 2 + j * \frac{D}{M}}, \dots, x_{i, (j + 1) * \frac{D}{M}})  &&
        \end{align*}\;
        \nonl \textbf{Chunk-Wise Learning Step:}  Create a dictionary $C_{j}$ with $K$ pivots using \textbf{\autoref{alg:chunk-wise-accurate-pq}}.\;
    }
    \textbf{Encoding Step:} Create a codebook $\mathcal{C} = \{C_{1}, C_{2}, \dots, C_{M}\}$. For each chunk vector $x^{j}_{i}$, pick a pivot $\mu_{j} \in C_{j}$ according to the \textbf{Membership Assignment Step} of  \textbf{\autoref{alg:chunk-wise-accurate-pq}}. Hence, each base point $x_{i}$ can be encoded as a concatenation of $\{\mu_{1}, \mu_{2}, \dots, \mu_{M}\}$.\;
}
\end{algorithm}

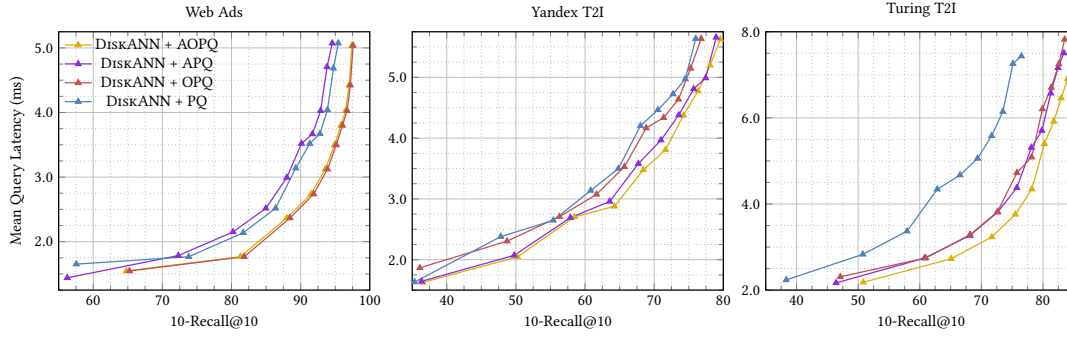
\begin{figure*}
\flushleft
\begin{minipage}[c]{\columnwidth}
    \begin{tikzpicture}[scale=0.6,font=\large]
    \begin{axis}[
        title=Web Ads,
        width=\linewidth,
        line width=0.8,
        grid=both,
        grid style={line width=.1pt, draw=gray!10},
        major grid style={line width=.2pt,draw=gray!50},
        minor grid style={line width=.3pt,draw=gray!100, dotted},
        minor tick num=3,
        tick label style={font=\large},
        legend style={nodes={scale=1.0, transform shape}},
        label style={font=\large},
        legend image post style={mark=triangle*},
        grid style={white},
        xlabel={10-Recall@10},
        ylabel={Mean Query Latency (ms)},
        y tick label style={
            /pgf/number format/.cd,
            fixed,
            fixed zerofill,
            precision=1
        },
        legend style={at={(0,1), font=\normalsize}, anchor=north west, draw=none, fill=none},
        legend columns=1,
        ymin=1.25,
        ymax=5.25,
        xmin=55,
        xmax=100,
    ]
        \addplot[yyellow, mark=triangle*] coordinates
          {(64.77,1.546) (81.27,1.767) (87.99,2.367) (91.50,2.738) (93.56,3.123) (94.89,3.497) (95.83,3.800) (96.51,4.029) (97.01,4.424) (97.41,5.036)};
          \addlegendentry{\previoustool + AOPQ}

        \addplot[ppurple, mark=triangle*] coordinates
          {(56.24,1.440) (72.31,1.785) (80.22,2.151) (84.96,2.516) (88.00,2.993) (90.12,3.519) (91.70,3.667) (92.89,4.029) (93.82,4.707) (94.57,5.073)};
          \addlegendentry{\previoustool + APQ}

        \addplot[rred, mark=triangle*] coordinates
          {(65.24,1.546) (81.84,1.767) (88.46,2.367) (91.88,2.738) (93.88,3.123) (95.15,3.497) (96.05,3.800) (96.72,4.029) (97.19,4.424) (97.58,5.036)};
          \addlegendentry{\previoustool + OPQ}
          
        \addplot[bblue, mark=triangle*] coordinates
          {(57.55,1.652) (73.81,1.767) (81.69,2.138) (86.36,2.516) (89.29,3.138) (91.32,3.519) (92.78,3.671) (93.88,4.042) (94.74,4.685) (95.44,5.073)};
          \addlegendentry{\previoustool + PQ}
    \end{axis}
    \hspace{4.7cm}
    \begin{axis}[
        title=Yandex T2I,
        width=\linewidth,
        line width=0.8,
        grid=both,
        grid style={line width=.1pt, draw=gray!10},
        major grid style={line width=.2pt,draw=gray!50},
        minor grid style={line width=.3pt,draw=gray!100, dotted},
        minor tick num=3,
        tick label style={font=\large},
        legend style={nodes={scale=1.0, transform shape}},
        label style={font=\large},
        legend image post style={mark=triangle*},
        grid style={white},
        xlabel={10-Recall@10},
        y tick label style={
            /pgf/number format/.cd,
            fixed,
            fixed zerofill,
            precision=1
        },
        legend style={at={(0,1), font=\normalsize}, anchor=north west, draw=none, fill=none},
        legend columns=1,
        ymin=1.5,
        ymax=5.75,
        xmin=35,
        xmax=80,
    ]
        \addplot[yyellow, mark=triangle*] coordinates
          {(36.51,1.626) (50.22,2.047) (58.47,2.704) (64.25,2.878) (68.43,3.478) (71.65,3.810) (74.23,4.374) (76.35,4.775) (78.08,5.198) (79.58,5.627)};

        \addplot[ppurple, mark=triangle*] coordinates
          {(36.35,1.645) (49.75,2.072) (57.89,2.692) (63.56,2.958) (67.71,3.580) (70.98,3.967) (73.56,4.380) (75.71,4.810) (77.45,4.989) (78.96,5.658)};

        \addplot[rred, mark=triangle*] coordinates
          {(36.10,1.866) (48.71,2.305) (56.30,2.708) (61.67,3.075) (65.69,3.529) (68.84,4.167) (71.36,4.336) (73.51,4.638) (75.27,5.148) (76.79,5.637)};
          
        \addplot[bblue, mark=triangle*] coordinates
          {(35.35,1.633) (47.81,2.380) (55.40,2.647) (60.82,3.138) (64.82,3.499) (68.01,4.202) (70.54,4.465) (72.72,4.725) (74.49,4.972) (76.01,5.637)};
    \end{axis}
    \hspace{4.7cm}
    \begin{axis}[
        title=Turing T2I,
        width=\linewidth,
        line width=0.8,
        grid=both,
        grid style={line width=.1pt, draw=gray!10},
        major grid style={line width=.2pt,draw=gray!50},
        minor grid style={line width=.3pt,draw=gray!100, dotted},
        minor tick num=3,
        tick label style={font=\large},
        legend style={nodes={scale=1.0, transform shape}},
        label style={font=\large},
        legend image post style={mark=triangle*},
        grid style={white},
        xlabel={10-Recall@10},
        y tick label style={
            /pgf/number format/.cd,
            fixed,
            fixed zerofill,
            precision=1
        },
        legend style={at={(0,1), font=\normalsize}, anchor=north west, draw=none, fill=none},
        legend columns=1,
        ymin=2.0,
        ymax=8.0,
        xmin=35,
        xmax=85.5,
    ]
        \addplot[yyellow, mark=triangle*] coordinates
          {(50.84,2.182) (65.10,2.729) (71.69,3.233) (75.52,3.759) (78.19,4.345) (80.23,5.398) (81.75,5.917) (82.96,6.459) (83.98,6.908) (84.85,7.751)};

        \addplot[ppurple, mark=triangle*] coordinates
          {(46.42,2.168) (60.78,2.743) (68.11,3.266) (72.63,3.810) (75.76,4.376) (78.11,5.310) (79.83,5.706) (81.27,6.572) (82.44,7.164) (83.39,7.505)};

        \addplot[rred, mark=triangle*] coordinates
          {(47.10,2.309) (60.97,2.746) (68.17,3.287) (72.63,3.823) (75.78,4.727) (78.16,5.087) (79.91,6.206) (81.36,6.705) (82.56,7.248) (83.52,7.822)};
          
        \addplot[bblue, mark=triangle*] coordinates
          {(38.33,2.237) (50.76,2.831) (57.93,3.368) (62.84,4.342) (66.54,4.672) (69.38,5.058) (71.67,5.584) (73.49,6.146) (75.13,7.260) (76.50,7.436)};
    \end{axis}
\end{tikzpicture}
\end{minipage}\hfill
\begin{minipage}[c]{0.4\columnwidth}
    \caption{(Left to right) Latency vs.\ Recall for AOPQ, APQ, OPQ and PQ at 10M scale.}
    \label{fig:combined-apq}
\end{minipage}
\vspace{-5pt}
\end{figure*}


\subsection{Improvement Over PQ and OPQ}
As evident from \textbf{\autoref{fig:combined-apq}}, we observe a $3\text{-}4\%$ improvement in recall, for the same latency target, in the case of text-to-image datasets. We also observe a negligible change in performance for the minimally OOD Web Ads queries. Since we need to construct query sets for each base point, we have a disk space overhead of roughly $400$ bytes per base point (for storing roughly $100$ query ids). Additionally, we train Accurate PQ through gradient descent, which takes roughly $1$ minute per chunk using Tensorflow~\citep{TF} on a $32$-core CPU.

\section{\parallelgorder Graph Reordering}
\label{sec:graph-reordering}

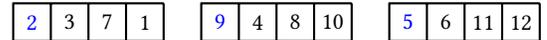
\begin{figure}
    \begin{tikzpicture}

        \tikzset{vertex/.style = {shape=circle,draw,minimum size=2.0em}}
        \tikzset{edge/.style = {->,> = latex'}}
        \node[vertex, thick] (1) at  (1.5,0) {$1$};
        \node[vertex, thick] (2) at  (2.3,1.2) {$2$};
        \node[vertex, thick] (3) at  (3.5,1.2) {$3$};
        \node[vertex, thick] (4) at  (4.3,0) {$4$};
        \node[vertex, thick] (5) at  (2.3,-1.2) {$5$};
        \node[vertex, thick] (6) at  (3.5,-1.2) {$6$};
        \node[vertex, thick] (7) at  (2.9,0) {$7$};
        \node[vertex, thick] (8) at  (5.3,-1.2) {$8$};
        \node[vertex, thick] (9) at  (5.5,0) {$9$};
        \node[vertex, thick] (10) at  (5.3,1.2) {$10$};
        \node[vertex, thick] (11) at  (6.5,1) {$11$};
        \node[vertex, thick] (12) at  (6.5,-1) {$12$};
    
        \draw[edge, thick] (7) to (1);
        \draw[edge, thick] (7) to (2);
        \draw[edge, thick] (2) to (3);
        \draw[edge, thick] (7) to (3);
        \draw[edge, thick] (7) to (4);
        \draw[edge, thick] (7) to (5);
        \draw[edge, thick] (5) to (6);
        \draw[edge, thick] (7) to (6);
        \draw[edge, thick] (4) to[bend left] (9);
        \draw[edge, thick] (9) to[bend left] (4);
        \draw[edge, thick] (9) to (10);
        \draw[edge, thick] (9) to (11);
        \draw[edge, thick] (9) to (8);
        \draw[edge, thick] (9) to (12);
        
        \draw[thick,color=black,xstep=0.5cm,ystep=0.5cm] (0.99,-2.51) grid (7,-2.0);
        \node[text=blue] at (1.25,-2.25) {$2$};
        \node at (1.75,-2.25) {$3$};
        \node at (2.25,-2.25) {$7$};
        \node at (2.75,-2.25) {$1$};
        \node at (3.25,-2.25) {$4$};
        \node at (3.75,-2.25) {$5$};
        \node at (4.25,-2.25) {$6$};
        \node at (4.75,-2.25) {$9$};
        \node at (5.25,-2.25) {$8$};
        \node at (5.75,-2.25) {$10$};
        \node at (6.25,-2.25) {$11$};
        \node at (6.75,-2.25) {$12$};
        \node [below=0.25cm, align=flush center, text width=8cm] at (4, -2.5){$\text{ \gorder Ordering}$ for the above graph.};
        
        \draw[thick,color=black,xstep=0.5cm,ystep=0.5cm] (0.49,-4.0) grid (2.5,-3.5);
        \draw[thick,color=black,xstep=0.5cm,ystep=0.5cm] (2.99,-4.0) grid (5,-3.5);
        \draw[thick,color=black,xstep=0.5cm,ystep=0.5cm] (5.49,-4.0) grid (7.5,-3.5);
        \node[text=blue] at (0.75,-3.75) {$2$};
        \node at (1.25,-3.75) {$3$};
        \node at (1.75,-3.75) {$7$};
        \node at (2.25,-3.75) {$1$};
        \node[text=blue] at (3.25,-3.75) {$9$};
        \node at (3.75,-3.75) {$4$};
        \node at (4.25,-3.75) {$8$};
        \node at (4.75,-3.75) {$10$};
        \node[text=blue] at (5.75,-3.75) {$5$};
        \node at (6.25,-3.75) {$6$};
        \node at (6.75,-3.75) {$11$};
        \node at (7.25,-3.75) {$12$};
        \node [below=0.25cm, align=flush center, text width=8cm] at (4, -4.0){ A possible $\text{\parallelgorder Ordering.}$};

    \end{tikzpicture}
    \caption{Graph ordering example for a window of size 4. Seed nodes are highlighted in blue. Each window signifies a packed disk sector.}
    \label{fig:ordering_example}
    \vspace{-10pt}
\end{figure}
\begin{algorithm}
\caption{\parallelgorder($S_{\text{sector}}$, $S_{\text{node}}$)}
\label{alg:parallel-gorder}
\DontPrintSemicolon
\small
\SetKwFor{ParallelFor}{parallel for}{do}{endfor}
\SetKwFunction{FSectorPack}{SectorPack}
\KwData{Sector size $S_{\text{sector}}$ and max node size $S_{\text{node}}$ (both in bytes).}
\KwResult{Array $P$ holding the new order of nodes.}
\Begin{
    \nonl $\text{Initialize empty array } P \text{ of size } |\mathcal{X}|.$\;
    \nonl $w \leftarrow \lfloor\frac{S_{\text{sector}}}{S_{\text{node}}}\rfloor ; \mathcal{D} \leftarrow \{v \mapsto \text{false}\}\text{ } \forall \text{ } v \in \mathcal{X}$\;
    \ParallelFor{$i \in [0, 1, \dots, \lfloor\frac{|\mathcal{X}|}{w} \rfloor - 1]$}{
        \nonl $\text{Pick a random, unpacked seed node } s.$\;
        \nonl \FSectorPack{$P[i * w]$, $\mathcal{D}$, $s$, $w$,}\;
    }
    \nonl $\text{Pick a random, unpacked seed node } s.$\;
    \nonl \FSectorPack{$P[\lfloor\frac{|\mathcal{X}|}{w} \rfloor * w]$, $\mathcal{D}$, $s$, $w$, }\;
}
\end{algorithm}
\begin{algorithm}
\caption{\sectorpack($P$, $\mathcal{D}$, $s$, $w$)}
\label{alg:sector-pack}
\DontPrintSemicolon
\small
\SetKwFunction{FIncKey}{IncrementKey}
\SetKwFunction{FTop}{top}
\SetKwFunction{FPop}{pop}
\SetKwFunction{FEmpty}{empty}
\SetKwFunction{FKeys}{keys}
\SetKwFunction{FGet}{get\_count}
\SetKwFunction{FInsert}{insert}
\SetKwFunction{FDelete}{delete}
\KwData{Sub-array $P$, bitmap $\mathcal{D}$ of already packed nodes, seed node $s$ and the number of elements to pack $w$.}
\KwResult{Sub-array $P$ with nodes to pack in the disk sector.}

\SetKwProg{myFIncKey}{Member Function}{}{}
\myFIncKey{$\mathcal{H}::\hspace{2pt}$\FIncKey{$v$}}{
    \eIf{$v \in \mathcal{H}$.\FKeys{}}{
        \nonl $v_{\text{count}} \leftarrow \mathcal{H}.$\FGet{$v$}\;
        \nonl $\mathcal{H}.$\FDelete{$v$} $; \mathcal{H}.$\FInsert{$v, v_{\text{count}} + 1$}\;
    }{
        \nonl $\mathcal{H}.$\FInsert{$v, 1$}\;
    }
}

\Begin{
    \nonl $\text{Initialize empty max heap } \mathcal{H}.$\;
    \nonl $i \leftarrow 0; P[i] \leftarrow s$\;
    \While{$i < w$}{
        \nonl $v_{e} \leftarrow P[i]; i \leftarrow i + 1$\;
        \For{$u \in N_{\text{out}}(v_{e})$}{
            \nonl $\mathcal{H}$.\FIncKey{$u$}\;
        }
        \For{$u \in N_{\text{in}}(v_{e})$}{
            \nonl $\mathcal{H}$.\FIncKey{$u$}\;
            \For{$t \in N_{\text{out}}(u)$}{
                \nonl $\mathcal{H}$.\FIncKey{$t$}\;
            }
        }
        \While{{\normalfont true}}{
            \If{$\mathcal{H}$.\FEmpty{}}{
                \nonl $\text{Pick a random unpacked seed node } v_{\text{max}}.$\;
                \nonl $\textbf{break}$\;
            }
            \nonl $v_{\text{max}} \leftarrow \mathcal{H}$.\FTop{} $; \mathcal{H}$.\FPop{}\;
            \If{{\normalfont not} $ \mathcal{D}[v_{\text{max}}]$}{
                \nonl $\textbf{break}$\;
            }
        }
        \nonl $\mathcal{D}[v_{\text{max}}] \leftarrow \text{true}; P[i] \leftarrow v_{\text{max}}$\;
    }
}
\end{algorithm}

\begin{figure*}
\flushleft
\begin{minipage}[c]{\columnwidth}
    \begin{tikzpicture}[scale=0.6,font=\large]
    \begin{axis}[
        title=Web Ads,
        width=\linewidth,
        line width=0.8,
        grid=both,
        grid style={line width=.1pt, draw=gray!10},
        major grid style={line width=.2pt,draw=gray!50},
        minor grid style={line width=.3pt,draw=gray!100, dotted},
        minor tick num=3,
        tick label style={font=\large},
        legend style={nodes={scale=1.0, transform shape}},
        label style={font=\large},
        legend image post style={mark=triangle*},
        grid style={white},
        xlabel={10-Recall@10},
        ylabel={Mean Query Latency (ms)},
        y tick label style={
            /pgf/number format/.cd,
            fixed,
            fixed zerofill,
            precision=1
        },
        legend style={at={(0,1), font=\normalsize}, anchor=north west, draw=none, fill=none},
        legend columns=1,
        ymin=0.75,
        ymax=4.5,
        xmin=55,
        xmax=97.5,
    ]
        \addplot[bblue, mark=triangle*] coordinates
          {(61.77,1.034) (77.50,1.370) (84.38,1.731) (88.29,2.073) (90.76,2.431) (92.48,2.797) (93.71,3.164) (94.66,3.515) (95.40,3.888) (96.00,4.249)};
          \addlegendentry{\previoustool + Random Order (ID)}

        \addplot[rred, mark=triangle*] coordinates
          {(61.79,0.886) (77.51,1.143) (84.39,1.393) (88.29,1.647) (90.77,1.912) (92.49,2.150) (93.72,2.429) (94.67,2.678) (95.41,2.951) (96.00,3.172)};
         \addlegendentry{\previoustool + \parallelgorder (ID)}

        \addplot[yyellow, mark=triangle*] coordinates
          {(57.68,1.061) (74.00,1.392) (81.85,1.740) (86.44,2.082) (89.33,2.461) (91.35,2.823) (92.84,3.182) (93.95,3.531) (94.81,3.908) (95.48,4.253)};
          \addlegendentry{\previoustool + Random Order (OOD)}

        \addplot[ppurple, mark=triangle*] coordinates
          {(57.72,0.927) (74.02,1.187) (81.85,1.446) (86.44,1.717) (89.32,1.980) (91.36,2.241) (92.84,2.495) (93.95,2.741) (94.81,3.009) (95.48,3.283)};
         \addlegendentry{\previoustool + \parallelgorder (OOD)}
    \end{axis}
    \hspace{4.7cm}
    \begin{axis}[
        title=Yandex T2I,
        width=\linewidth,
        line width=0.8,
        grid=both,
        grid style={line width=.1pt, draw=gray!10},
        major grid style={line width=.2pt,draw=gray!50},
        minor grid style={line width=.3pt,draw=gray!100, dotted},
        minor tick num=3,
        tick label style={font=\large},
        legend style={nodes={scale=1.0, transform shape}},
        label style={font=\large},
        legend image post style={mark=triangle*},
        grid style={white},
        xlabel={10-Recall@10},
        y tick label style={
            /pgf/number format/.cd,
            fixed,
            fixed zerofill,
            precision=1
        },
        legend style={at={(0,1), font=\normalsize}, anchor=north west, draw=none, fill=none},
        legend columns=1,
        ymin=0.75,
        ymax=5.25,
        xmin=35,
        xmax=100,
    ]
        \addplot[bblue, mark=triangle*] coordinates
          {(73.47,1.092) (89.31,1.434) (94.40,1.798) (96.48,2.287) (97.45,2.562) (97.98,3.123) (98.29,3.332) (98.49,3.706) (98.63,4.116) (98.73,4.463)};

        \addplot[rred, mark=triangle*] coordinates
          {(73.46,0.978) (89.31,1.245) (94.40,1.542) (96.48,1.835) (97.45,2.124) (97.98,2.401) (98.29,2.704) (98.49,2.978) (98.63,3.578) (98.73,3.708)};

        \addplot[yyellow, mark=triangle*] coordinates
          {(35.60,1.295) (47.95,1.965) (55.49,2.107) (60.81,2.488) (64.81,2.877) (68.01,3.268) (70.57,3.884) (72.70,4.050) (74.49,4.704) (76.05,5.074)};

        \addplot[ppurple, mark=triangle*] coordinates
          {(35.61,1.185) (47.95,1.757) (55.48,1.867) (60.81,2.182) (64.80,2.509) (68.01,2.841) (70.57,3.177) (72.70,3.471) (74.49,3.818) (76.05,4.139)};
    \end{axis}
    \hspace{4.7cm}
    \begin{axis}[
        title=Turing T2I,
        width=\linewidth,
        line width=0.8,
        grid=both,
        grid style={line width=.1pt, draw=gray!10},
        major grid style={line width=.2pt,draw=gray!50},
        minor grid style={line width=.3pt,draw=gray!100, dotted},
        minor tick num=3,
        tick label style={font=\large},
        legend style={nodes={scale=1.0, transform shape}},
        label style={font=\large},
        legend image post style={mark=triangle*},
        grid style={white},
        xlabel={10-Recall@10},
        y tick label style={
            /pgf/number format/.cd,
            fixed,
            fixed zerofill,
            precision=1
        },
        legend style={at={(0,1), font=\normalsize}, anchor=north west, draw=none, fill=none},
        legend columns=1,
        ymin=1.5,
        ymax=7.5,
        xmin=35,
        xmax=100,
    ]
        \addplot[bblue, mark=triangle*] coordinates
          {(86.20,1.753) (96.21,2.165) (97.70,2.754) (98.31,3.242) (98.65,3.803) (98.85,4.340) (99.00,4.804) (99.10,5.337) (99.18,6.013) (99.23,6.503)};

        \addplot[rred, mark=triangle*] coordinates
          {(86.19,1.641) (96.21,2.005) (97.69,2.440) (98.31,2.843) (98.64,3.256) (98.85,3.579) (98.99,4.072) (99.10,4.516) (99.17,5.106) (99.22,5.477)};

        \addplot[yyellow, mark=triangle*] coordinates
          {(38.26,2.042) (50.57,2.621) (57.74,3.187) (62.69,3.690) (66.34,4.286) (69.20,4.796) (71.40,5.360) (73.34,6.041) (74.92,6.487) (76.33,6.981)};

        \addplot[ppurple, mark=triangle*] coordinates
          {(38.27,1.923) (50.58,2.439) (57.74,2.942) (62.70,3.367) (66.36,3.893) (69.21,4.363) (71.42,4.764) (73.35,5.377) (74.93,5.884) (76.33,6.264)};
    \end{axis}
\end{tikzpicture}
\end{minipage}\hfill
\begin{minipage}[c]{0.4\columnwidth}
    \caption{(Left to right) Latency vs.\ Recall for Random Order and \parallelgorder at 10M scale.}
    \label{fig:combined-reorder-io-latency}
\end{minipage}
\vspace{-5pt}
\end{figure*}
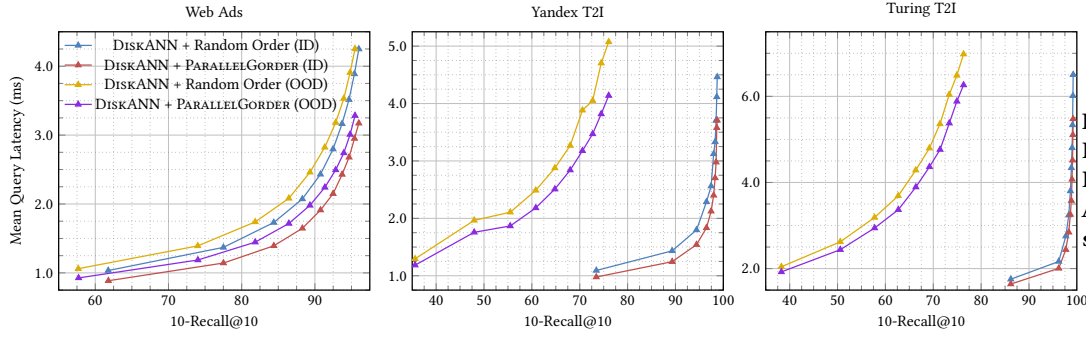

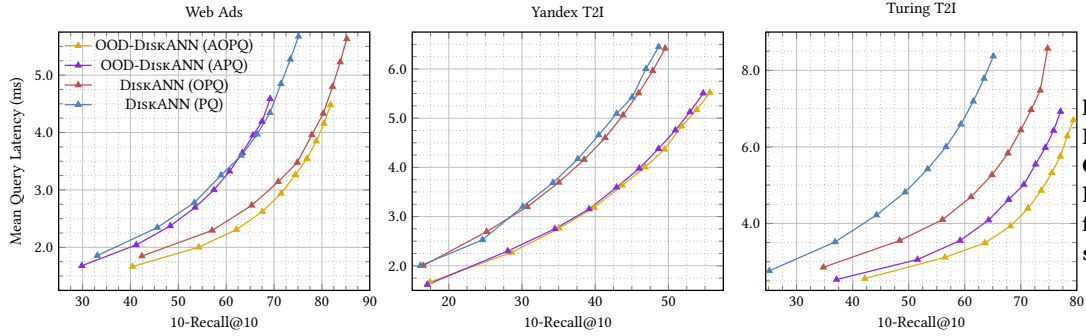
\begin{figure*}[ht]
\flushleft
\begin{minipage}[c]{\columnwidth}
    \begin{tikzpicture}[scale=0.6,font=\large]
    \begin{axis}[
        title=Web Ads,
        width=\linewidth,
        line width=0.8,
        grid=both,
        grid style={line width=.1pt, draw=gray!10},
        major grid style={line width=.2pt,draw=gray!50},
        minor grid style={line width=.3pt,draw=gray!100, dotted},
        minor tick num=3,
        tick label style={font=\large},
        legend style={nodes={scale=1.0, transform shape}},
        label style={font=\large},
        legend image post style={mark=triangle*},
        grid style={white},
        xlabel={10-Recall@10},
        ylabel={Mean Query Latency (ms)},
        y tick label style={
            /pgf/number format/.cd,
            fixed,
            fixed zerofill,
            precision=1
        },
        legend style={at={(0,1), font=\normalsize}, anchor=north west, draw=none, fill=none},
        legend columns=1,
        ymin=1.25,
        ymax=5.75,
        xmin=25,
        xmax=90.0,
    ]
        \addplot[yyellow, mark=triangle*] coordinates
        {(40.44,1.666) (54.27,2.003) (62.14,2.309) (67.55,2.622) (71.47,2.939) (74.47,3.265) (76.87,3.544) (78.80,3.849) (80.44,4.157) (81.80,4.482)};
          \addlegendentry{\tool (AOPQ)}

        \addplot[ppurple, mark=triangle*] coordinates
        {(29.83,1.679) (41.22,2.043) (48.32,2.373) (53.51,2.696) (57.45,3.000) (60.70,3.324) (63.36,3.645) (65.61,3.953) (67.49,4.191) (69.18,4.589)};
          \addlegendentry{\tool (APQ)}

        \addplot[rred, mark=triangle*] coordinates
        {(42.39,1.848) (57.07,2.295) (65.33,2.731) (70.87,3.142) (74.84,3.477) (77.89,3.959) (80.28,4.332) (82.22,4.800) (83.82,5.227) (85.18,5.631)};
          \addlegendentry{\previoustool(OPQ)}
          
        \addplot[bblue, mark=triangle*] coordinates
        {(33.07,1.856) (45.61,2.345) (53.35,2.780) (58.91,3.255) (63.14,3.600) (66.47,3.969) (69.15,4.345) (71.45,4.850) (73.40,5.274) (75.11,5.675)};
          \addlegendentry{\previoustool (PQ)}
    \end{axis}
    \hspace{4.7cm}
    \begin{axis}[
        title=Yandex T2I,
        width=\linewidth,
        line width=0.8,
        grid=both,
        grid style={line width=.1pt, draw=gray!10},
        major grid style={line width=.2pt,draw=gray!50},
        minor grid style={line width=.3pt,draw=gray!100, dotted},
        minor tick num=3,
        tick label style={font=\large},
        legend style={nodes={scale=1.0, transform shape}},
        label style={font=\large},
        legend image post style={mark=triangle*},
        grid style={white},
        xlabel={10-Recall@10},
        y tick label style={
            /pgf/number format/.cd,
            fixed,
            fixed zerofill,
            precision=1
        },
        legend style={at={(0,1), font=\normalsize}, anchor=north west, draw=none, fill=none},
        legend columns=1,
        ymin=1.5,
        ymax=6.75,
        xmin=15,
        xmax=57.5,
    ]
        \addplot[yyellow, mark=triangle*] coordinates
          {(17.43,1.671) (28.56,2.272) (35.08,2.762) (39.86,3.184) (43.69,3.635) (46.83,4.005) (49.46,4.366) (51.81,4.836) (53.86,5.172) (55.67,5.521)};

        \addplot[ppurple, mark=triangle*] coordinates
          {(17.07,1.618) (28.06,2.300) (34.49,2.751) (39.18,3.154) (42.91,3.595) (46.02,3.978) (48.64,4.378) (50.95,4.755) (52.95,5.127) (54.76,5.507)};

        \addplot[rred, mark=triangle*] coordinates
          {(16.50,2.002) (25.17,2.694) (30.73,3.200) (35.03,3.695) (38.49,4.159) (41.36,4.600) (43.80,5.063) (45.93,5.510) (47.85,5.963) (49.56,6.418)};

        \addplot[bblue, mark=triangle*] coordinates
          {(16.13,2.002) (24.60,2.527) (30.10,3.203) (34.23,3.688) (37.63,4.172) (40.49,4.657) (42.92,5.092) (45.05,5.425) (46.92,6.002) (48.66,6.449)};
    \end{axis}
    \hspace{4.7cm}
    \begin{axis}[
        title=Turing T2I,
        width=\linewidth,
        line width=0.8,
        grid=both,
        grid style={line width=.1pt, draw=gray!10},
        major grid style={line width=.2pt,draw=gray!50},
        minor grid style={line width=.3pt,draw=gray!100, dotted},
        minor tick num=3,
        tick label style={font=\large},
        legend style={nodes={scale=1.0, transform shape}},
        label style={font=\large},
        legend image post style={mark=triangle*},
        grid style={white},
        xlabel={10-Recall@10},
        y tick label style={
            /pgf/number format/.cd,
            fixed,
            fixed zerofill,
            precision=1
        },
        legend style={at={(0,1), font=\normalsize}, anchor=north west, draw=none, fill=none},
        legend columns=1,
        ymin=2.25,
        ymax=9.0,
        xmin=24.5,
        xmax=80,
    ]
        \addplot[yyellow, mark=triangle*] coordinates
          {(42.16,2.559) (56.47,3.113) (63.60,3.491) (68.16,3.930) (71.27,4.391) (73.65,4.860) (75.52,5.324) (77.05,5.753) (78.30,6.282) (79.39,6.697)};

        \addplot[ppurple, mark=triangle*] coordinates
          {(37.13,2.534) (51.57,3.056) (59.17,3.552) (64.25,4.087) (67.82,4.622) (70.54,5.016) (72.64,5.546) (74.38,5.979) (75.85,6.428) (77.09,6.923)};

        \addplot[rred, mark=triangle*] coordinates
          {(34.78,2.853) (48.39,3.552) (56.03,4.099) (61.13,4.691) (64.86,5.270) (67.71,5.833) (70.00,6.441) (71.84,6.966) (73.46,7.477) (74.78,8.572)};
          
        \addplot[bblue, mark=triangle*] coordinates
          {(25.14,2.761) (36.87,3.515) (44.25,4.218) (49.37,4.818) (53.38,5.420) (56.65,5.999) (59.31,6.590) (61.52,7.188) (63.43,7.786) (65.10,8.368)};
    \end{axis}
\end{tikzpicture}
\end{minipage}\hfill
\begin{minipage}[c]{0.4\columnwidth}
    \caption{(Left to right) Latency vs.\ Recall for \tool and \previoustool with different quantization schemes at full scale.}
    \label{fig:combined-1b}
\end{minipage}
\vspace{-5pt}
\end{figure*}

Next, we show how we can \emph{re-order} the graph nodes to improve locality of reference during graph traversal for index search. While there has been some recent work on this topic for million-scale and memory-resident indices such as \hnsw~\cite{coleman-survey}, we present \parallelgorder with two notable differences: (a) it is highly parallelizable, enabling it to re-order billion-scale graphs, and (b) it optimizes for the number of SSD I/Os during search, for disk-resident graphs, as opposed to the cache miss rate considered in prior work.

Graph-based ANNS algorithms, at their core, traverse nodes in the graph in a greedy manner by walking over to the closest, unexplored node to the query in the candidate list. After arriving at a particular node, we need to access the data vectors of its out-neighbors. These accesses are essentially \emph{random accesses}, leading to large cache miss rates. Moreover a node, once explored, is never referenced again for that query. This essentially means that greedy graph traversal has poor inherent locality of reference.

Recently, there has been some work on \emph{graph reordering methods} to improve search latency by placing information regarding nodes likely to be referenced together on the same or adjacent cache lines. This has the effect of reducing the number of cache misses incurred during graph traversal, as one needs fewer cache line transfers to access the set of referenced nodes. 
\citet{Gorder} introduces a greedy heuristic called \gorder to compute such a re-ordering, and apply it to the \hnsw algorithm for small, million-scale indices. An interested reader may also refer to \citet{coleman-survey}, which has surveyed several other reordering techniques applied to the \hnsw algorithm, and consistently observed the best speedups with \gorder algorithm for in-memory graph ANNS.

\subsection{Locality for External Memory}
While previous work on re-ordering has focused on small-scale RAM-resident graph indices, these in-memory indices become prohibitively expensive for large-scale datasets.

We argue that there is an equally strong motivation for re-ordering on billion-scale indices as well. As the \previoustool greedy search traverses a graph, the out-neighborhood of the referenced node is fetched by making a random SSD read request. One of the major bottlenecks for search performance on such disk-based indices is the maximum number of random I/O Operations Per Second (IOPS) supported by the SSD. Waiting on such I/O typically accounts for more than $50\%$ of the overall latency for \previoustool even when used with the latest NVMe (Non-Volatile Memory Express) SSDs. Our main idea is that, since random reads on SSDs are typically done in sectors of $4$KBs or larger, there is scope for reducing the number of SSD reads if we optimize the index layout by \emph{placing nodes which are likely to be referenced in the same search path, in the same SSD sector}.

While \gorder has shown reasonable success for small datasets, there are significant scalability issues for larger datasets ($> 100$ million). Vanilla \gorder is ill-suited for this task, since it has an $\mathcal{O}(NR^{2})$ time complexity of computing the re-ordering, where $N$ denotes the number of graph nodes, and $R$ denotes the maximum out-degree. Moreover, the algorithm is based on a sliding window-based greedy implementation, and is inherently non-parallelizable. As a result, the total build-time cost from our experiments ends up being around $2\times$ as much as the index construction. 

To this end, we make use of the fact that \emph{one only needs to maximise the locality between all the graph nodes stored in a particular disk sector}, without worrying about the locality of nodes spread across two different sectors. This then opens up the possibility to ``pack'' each sector with nodes individually (see \textbf{\autoref{fig:ordering_example}}), \emph{making the entire process parallelizable}. In case of a race condition, where a node has multiple different sectors where it can be packed into, we simply choose a sector at random for the node, and re-initiate the packing process on the rest of the sectors.

\textbf{\autoref{alg:parallel-gorder}} and \textbf{\autoref{alg:sector-pack}} describe our approach, where we make use of a standard max heap for tracking the most suitable node to add to a partially-packed sector, starting with a randomly-chosen seed node. Note that using a max heap pushes the runtime of our algorithm to $\mathcal{O}(NR^{2}\log{R})$, which is only slightly worse than the original, non-parallel algorithm since $R$ is typically $< 150$.


\subsection{Observed Improvement}
We observe anywhere from $15\text{-}40\%$ reduction in I/O requests in our experiments. As evident from \textbf{\autoref{fig:combined-reorder-io-latency}}, this leads to $10\text{-}25\%$ improvement in mean latency, for the same recall target. These gains are obtained with roughly a $7\text{-}12\%$ increase in build-time, keeping the peak RAM and disk usage unchanged during both the build and search, across all datasets.

\section{Evaluation}
\label{sec:evaluation}


\textbf{Datasets:} We evaluate our work on three large-scale datasets. Two of these are Text-to-Image (T2I) datasets, namely Yandex Text-to-Image-1B (200 dimensions) and Turing Text-to-Image (1024 dimensions). 
In Yandex Text-to-Image-1B, the base  dataset consists of 1 billion image embeddings produced by the Se-ResNext-101~\citep{Se-ResNext-101} model, and queries are textual embeddings produced by a variant of the DSSM~\citep{DSSM} model. 
In Turing Text-to-Image, the base dataset consists of 87 million image embeddings, and the queries are textual embeddings, both generated by the Turing Bletchley~\citep{T-Bletchley} model.
The third, Web Ads, is an web query-to-advertisements dataset in $64$ dimensions. The base embeddings encode both the text and images in $2.4$ billion product advertisements scraped from a well-known web index, and encoded by a variant of the CLSM~\citep{RichCDSSM} model. 
For Yandex Text-to-Image-1B and Web Ads datasets, we evaluate our results on a set of $100$K test queries and for Turing Text-to-Image, we evaluate our results on a set of $30$K test queries. 
All queries used during build or evaluation, whether ID or OOD, are sampled randomly and without replacement from these datasets.

\textbf{Hardware:} All experiments were conducted on a bare-metal server with 2$\times$ Xeon Gold 6140 CPUs (36 cores, 72 threads), with $500$ GBs of DDR4 RAM and a 3.2TB Samsung PM1725a PCIe SSD.

\textbf{Parameters:} We initiated all of our experiments with $R = 64$, $L = 128$, $\alpha_{1} = 1.0$, $\alpha_{2} = 1.2$. For all PQ-based quantization, we made use of $8$-bit encoding ($256$ centroids per chunk) and use $T'=10$, $T=80000$ and $\phi=100$. We train the APQ/AOPQ pivots on a sample of $100$K base and $100$K query points. For RAM-resident compressed vectors, we set the number of chunks $M = \frac{D}{4}$, where $D$ was the dimensionality of the vectors of a given dataset. For the disk-resident quantized vectors, we used $M = D$. The build-time RAM budget was software limited to $200$ GBs. During search, we set a beam width of $4$, and set the number of search threads to $16$.

\textbf{Comparison with Existing Compression Schemes:} The comparison of our work is limited to PQ and OPQ because these two techniques have the lowest search-time compute footprints. As observed from Fig. 2 and 3 in ~\citet{LSQ-v2} and Fig. 4 in ~\citet{mathijs-decoder}, PQ and OPQ are about $10\times\text{-}100\times$ faster than the other approaches, making them suitable for low-latency disk-based ANNS. APQ and AOPQ have the same search-time compute footprints as PQ and OPQ respectively.
Recently introduced \faketool~\citep{SCANN} also offers the same compute requirements as PQ, however further analysis~\citep{SCANN-Rebuttal1} shows that the reported accuracy gains have been due to its usage of $4$-bit encoding ($16$ centroids per chunk), which vanish when compared against a $4$-bit PQ baseline~\citep{SCANN-Rebuttal2}.

\textbf{\autoref{fig:combined-1b}} demonstrates the contribution of all three approaches towards improving the accuracy vs.\ latency boundary of the large scale ANNS. Taken together, these techniques yield upto $40\%$ improvement in latency, or $15\%$ improvement in recall.

\section{Conclusion}
\label{sec:conclusion}
We have presented and evaluated a new framework for OOD ANNS. We extended the graph construction algorithm of \previoustool, making it suitable for search with OOD queries, while maintaining ID query search qualities. We investigated how current Product Quantization schemes are inefficient in handling the nuances of OOD ANNS, and provided an improved formulation which leads to better performance. We also parallelized a state-of-the-art graph reordering algorithm, and adapted it to disk-based ANNS scenario. By combining all of our contributions, we have established a new state-of-the-art ANNS solution for the OOD setting, and verified our results on three large-scale datasets.

\begin{acks}
We would like to thank Ranajoy Sadhukhan, Neel Karia, Siddharth Gollapudi, Gopal Srinivasa and Nipun Kwatra for their helpful feedback and discussions.
We would also like to thank Kriti Aggarwal, Owais Khan Mohammed, Xiaochuan Ni, Subhojit Som for providing the Turing Text-to-Image dataset.
\end{acks}

\bibliographystyle{ACM-Reference-Format}
\bibliography{sample-base}

\clearpage
\appendix
\section{Appendix}
\label{sec:appendix}

\subsection{ANNS Preliminaries}
\label{sec:anns-preliminaries}
ANNS algorithms typically work in two phases, namely the build and the search phase. During build phase, they construct a data structure (called index) over the input database (also called base dataset) of points and associated vectors. The index is then used for rapidly sifting through the database while looking for the nearest neighbors of a query vector, during the search phase.

Graph ANNS algorithms construct a graph index, where the edges constructed by the build algorithm are guided by the vector-to-vector distances between the base dataset of points. A graph node thus comprises a base point, its associated vector and the out-neighbors of the base point.


Existing graph algorithms are derived from four base classes of graphs~\citep{wang-vldb}: Delaunay Graphs (DGs), Minimum Spanning Trees (MSTs), K-Nearest Neighbor Graphs ($k$-NNGs) and Relative Neighborhood Graphs (RNGs). DGs are expensive to construct, by virtue of being almost fully connected in high dimensions, and hence do not find much practical use. MSTs guarantee node reachability, but at the cost of long traversal paths. $k$-NNGs, by definition, only construct short-ranged edges, and do not guarantee global connectivity. RNGs aim to balance the construction of short and long-ranged edges, and inspire the current state-of-the-art graph-based approaches such as \vamana and \hnsw.

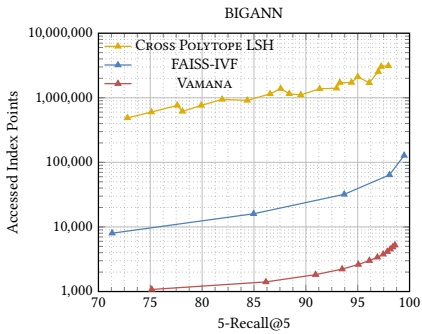
\begin{figure}[ht]
\centering
\vspace{-5pt}
\begin{minipage}{\columnwidth}
    \centering
    \begin{tikzpicture}[scale=0.6,font=\large]
    \begin{semilogyaxis}[
        title=BIGANN,
        width=\linewidth,
        line width=0.8,
        grid=both,
        grid style={line width=.1pt, draw=gray!10},
        major grid style={line width=.2pt,draw=gray!50},
        minor grid style={line width=.3pt,draw=gray!100, dotted},
        minor tick num=3,
        tick label style={font=\large},
        legend style={nodes={scale=1.0, transform shape}},
        label style={font=\large},
        legend image post style={mark=triangle*},
        grid style={white},
        xlabel={5-Recall@5},
        ylabel={Accessed Index Points},
        y tick label style={
            /pgf/number format/.cd,
            fixed,
            fixed zerofill,
            precision=1
        },
        legend style={at={(0,1), font=\normalsize}, anchor=north west, draw=none, fill=none},
        legend columns=1,
        ymin=1000,
        ymax=10000000,
        xmin=70,
        xmax=100,
    ]
        \addplot[yyellow, mark=triangle*] coordinates
          {(72.8,488701.8) (75.11,601012.7) (77.61,762975.9) (78.1,615321.5) (79.92,763380.9) (81.93,943559.4) (84.35,910992.3) (86.55,1150119.1) (87.57,1386989.8) (88.4,1153998.4) (89.51,1106455.7) (91.33,1378761.3) (92.96,1404776.5) (93.28,1719539.4) (94.4,1724914.3) (95.01,2115158.9) (96.12,1703905.7) (96.98,2539770.9) (97.27,3047971.2) (97.96,3133357.7)};
          \addlegendentry{\lsh}

        \addplot[bblue, mark=triangle*] coordinates
          {(71.32,8000) (84.96,16000) (93.70,32000) (98.06,64000) (99.48,128000)};
          \addlegendentry{\faiss}

        \addplot[rred, mark=triangle*] coordinates
          {(75.14,1076.9) (86.15,1410.4) (90.94,1824.3) (93.5,2226.7) (95.06,2622.7) (96.12,3012.4) (96.9,3396.9) (97.46,3775.9) (97.85,4147.2) (98.14,4517.0) (98.38,4880.5) (98.59,5240.7)};
          \addlegendentry{\vamana}
    \end{semilogyaxis}
\end{tikzpicture}
\end{minipage}
\caption{Accessed Index Points vs.\ Recall for BIGANN dataset (100M scale).}
\label{fig:graph-ivf-lsh}
\vspace{-5pt}
\end{figure}

State-of-the-art graph-based indices~\citep{HNSW, NSG, DiskANN} use data-dependent index construction to achieve $10\times\text{-}100\times$ more query efficiency over data-agnostic methods like LSH~\citep{FALCONN, LSH}, as measured by the number of index points accessed to achieve a certain recall (\textbf{\autoref{fig:graph-ivf-lsh}}).

Scaling ANNS solutions to billion-scale datasets requires one to adapt the built index to be disk-resident, while supporting low-latency search, because of RAM constraints. As such, disk-based solutions load compressed representations of the dataset vectors in RAM for fast and approximate distance estimates during search. The original, higher precision dataset vectors are stored as a component of the graph nodes on the disk. Once the compressed vector search has shortlisted a bunch of candidate nearest neighbors of a query, the disk-resident graph nodes, pertaining to these candidates, are paged into RAM through I/O requests for accurately re-ranking the candidates with higher precision vectors.

\begin{figure*}
\flushleft
\begin{minipage}[c]{\columnwidth}
    \begin{tikzpicture}[scale=0.6,font=\large]
    \begin{axis}[
        title=Yandex T2I ID,
        width=\linewidth,
        line width=0.8,
        grid=both,
        grid style={line width=.1pt, draw=gray!10},
        major grid style={line width=.2pt,draw=gray!50},
        minor grid style={line width=.3pt,draw=gray!100, dotted},
        minor tick num=3,
        tick label style={font=\large},
        legend style={nodes={scale=1.0, transform shape}},
        label style={font=\large},
        legend image post style={mark=triangle*},
        grid style={white},
        xlabel={10-Recall@10},
        ylabel={Mean Query Latency (ms)},
        y tick label style={
            /pgf/number format/.cd,
            fixed,
            fixed zerofill,
            precision=1
        },
        legend style={at={(0,1), font=\normalsize}, anchor=north west, draw=none, fill=none},
        legend columns=1,
        ymin=0.5,
        ymax=2.0,
        xmin=60,
        xmax=100,
    ]
        \addplot[rred, mark=triangle*] coordinates
          {(62.62,0.850) (80.84,0.953) (86.57,1.048) (89.44,1.137) (91.17,1.225) (92.34,1.311) (93.16,1.400) (93.74,1.487) (94.24,1.570) (94.63,1.655)};
          \addlegendentry{Yandex T2I ID PQ Search}

        \addplot[bblue, mark=triangle*] coordinates
          {(83.23,0.677) (89.12,0.809) (91.45,0.922) (92.78,1.033) (93.66,1.142) (94.31,1.254) (94.80,1.361) (95.18,1.465) (95.47,1.574) (95.71,1.680)};
          \addlegendentry{Yandex T2I ID Floating-Point Search}
    \end{axis}
    \hspace{4.7cm}
    \begin{axis}[
        title=Yandex T2I OOD,
        width=\linewidth,
        line width=0.8,
        grid=both,
        grid style={line width=.1pt, draw=gray!10},
        major grid style={line width=.2pt,draw=gray!50},
        minor grid style={line width=.3pt,draw=gray!100, dotted},
        minor tick num=3,
        tick label style={font=\large},
        legend style={nodes={scale=1.0, transform shape}},
        label style={font=\large},
        legend image post style={mark=triangle*},
        grid style={white},
        xlabel={10-Recall@10},
        y tick label style={
            /pgf/number format/.cd,
            fixed,
            fixed zerofill,
            precision=1
        },
        legend style={at={(0,1), font=\normalsize}, anchor=north west, draw=none, fill=none},
        legend columns=1,
        ymin=0.5,
        ymax=2.5,
        xmin=25,
        xmax=90.5,
    ]
        \addplot[rred, mark=triangle*] coordinates
          {(26.93,0.921) (41.75,1.075) (50.74,1.212) (57.00,1.341) (61.66,1.481) (65.26,1.606) (68.16,1.712) (70.54,1.833) (72.62,1.952) (74.34,2.063)};
          \addlegendentry{Yandex T2I OOD PQ Search}

        \addplot[bblue, mark=triangle*] coordinates
          {(57.45,0.740) (70.44,0.928) (76.81,1.097) (80.67,1.254) (83.31,1.410) (85.30,1.556) (86.81,1.706) (87.97,1.850) (88.97,1.992) (89.81,2.132)};
          \addlegendentry{Yandex T2I OOD Floating-Point Search}
    \end{axis}
    \hspace{5cm}
    \begin{axis}[
        title=Yandex T2I,
        width=\linewidth,
        line width=0.8,
        grid=both,
        grid style={line width=.1pt, draw=gray!10},
        major grid style={line width=.2pt,draw=gray!50},
        minor grid style={line width=.3pt,draw=gray!100, dotted},
        minor tick num=3,
        tick label style={font=\large},
        legend style={nodes={scale=1.0, transform shape}},
        label style={font=\large},
        legend image post style={mark=triangle*},
        grid style={white},
        xlabel={Percentile},
        ylabel={10-NN to 1-NN Distance Gap},
        y tick label style={
            /pgf/number format/.cd,
            fixed,
            fixed zerofill,
            precision=1
        },
        legend style={at={(0,1), font=\normalsize},
        anchor=north west, draw=none, fill=none},
        legend columns=1,
        ymin=0.0,
        ymax=0.75,
        xmin=-0.5,
        xmax=100.5,
    ]
        \addplot[rred, mark=triangle*] coordinates
          {(0,0.0021) (1,0.0136) (5,0.0260) (25,0.0534) (50,0.0782) (75,0.1110) (95,0.1869) (99,0.2875) (100,0.7416)};
          \addlegendentry{ID}

        \addplot[bblue, mark=triangle*] coordinates
          {(0,0.0023) (1,0.0113) (5,0.0184) (25,0.0328) (50,0.0469) (75,0.0661) (95,0.1070) (99,0.1491) (100,0.4076)};
          \addlegendentry{OOD}
    \end{axis}
\end{tikzpicture}
\end{minipage}\hfill
\begin{minipage}[c]{0.36\columnwidth}
    \caption{Latency vs.\ Recall for Yandex T2I on ID and OOD queries at 10M scale.}
    \label{fig:yandex-10m-pq-loss}
\end{minipage}
\vspace{-5pt}
\end{figure*}
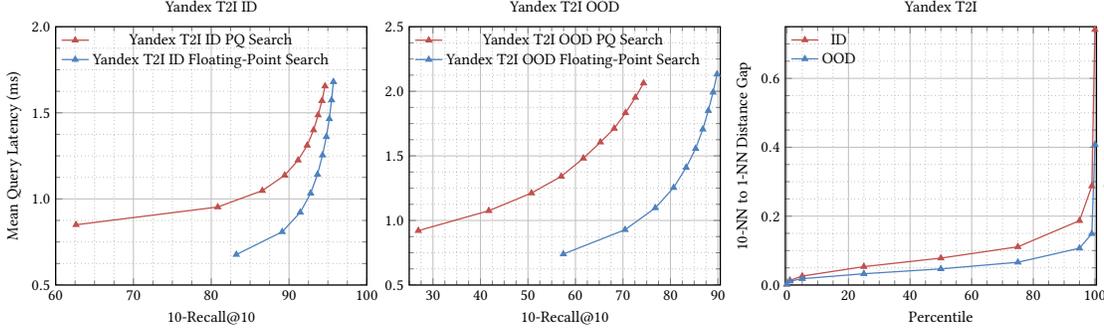

\subsection{Compression with Product Quantization}
\label{sec:pq-preliminaries}
Let $\mathcal{X} = \{x_{i} \in \mathbb{R}^{D} | i = 1, \dots, N\}$ be the set of $N$ base dataset vectors that we wish to compress, and $\mathcal{C} = \{\mu_{i} \in \mathbb{R}^{D} | i = 1, \dots, K\}$ be a set of $K$ learnable vectors (called pivots).

Quantization, in a classical sense, aims to learn a mapping $\mathcal{S}: \mathcal{X} \rightarrow \mathcal{C}$ such that the overall reconstruction error is minimised:
\begin{equation}
\begin{aligned}
    \text{min} \sum_{i=1}^{N} \| x_{i} - \mathcal{S}(x_{i})\|^{2}_{2},\text{ }\mathcal{S}(x_{i}) \in \mathcal{C}
\end{aligned}
\end{equation}
where $\mathcal{S}(x_{i})$ is the pivot (out of $K$ pivots) closest to the $x_{i}$.

This mapping is typically learnt through the $K$-Means clustering algorithm, where the cluster centroids serve as pivots, and are generated using an alternating minimization approach. We can then use the logarithm of the index values of the $K$ pivots, to assign bit-codes to individual base vectors for compression. The length of the bit codes is, thus, logarithmic in the number of pivots ($\log_2 K$).

The computational complexity of this approach is linear in $K$. Hence, generating a large number of pivots, which is desirable for minimising the reconstruction error, is computationally expensive through naive $K$-Means. This prohibits the generation of compressed codes with larger bit-lengths.

In Product Quantization (PQ), we can divide each base vector into $M$ concatenated sub-vectors (commonly referred to as ``chunks''), each of dimension $\frac{D}{M}$, and quantize each chunk with $K$ unique pivots. The classical formulation for PQ minimizes the following objective function:
\begin{equation}
\begin{aligned}
    \text{min}\sum_{j=1}^{M}\sum_{i=1}^{N}\| x^{j}_{i} - \mu\|^{2}_{2},\text{ }\mu \in \mathbb{R}^{\frac{D}{M}}
\end{aligned}
\end{equation}
where $x^{j}_{i}$ is the $j^{\text{th}}$ chunk of the $i^{\text{th}}$ vector, and $\mu$ is the closest pivot (out of the $K$ pivots in the $j^{\text{th}}$ chunk) to this sub-vector. Since each chunk has $K$ unique pivots, we have a total of $M \times K$ pivots which generate bit-codes of length $M \log_2 K$, whereas the naive technique would have required $K^{M}$ pivots to achieve the same bit-code length.

Optimized Product Quantization (OPQ) goes a step further, and learns an orthonormal matrix $\mathcal{R}$ for transforming the vector space of the dataset. It thus minimizes:
\begin{equation}
\begin{aligned}
    \text{min}\sum_{j=1}^{M}\sum_{i=1}^{N}\| (\mathcal{R}x_{i})^{j} - \mathcal{R}\mu\|^{2}_{2},\text{ }\mu \in \mathbb{R}^{\frac{D}{M}},\text{ }\mathcal{R}^{T}\mathcal{R}=\mathcal{I}
\end{aligned}
\end{equation}

\subsection{Why Do Existing PQ Schemes Fail In OOD Setting?}
\label{sec:pq-pitfalls}

Existing PQ schemes only consider the squared euclidean norm of the second multiplicative term as an upper-bound of the overall distance distortion. By triangle inequality, we have:
\begin{equation} \label{eq:9}
\begin{aligned}
    \underbrace{\|q^{j} - x^{j}_{i}\|_{2}}_{\text{Actual}} \leq \underbrace{\|q^{j} - \mu\|_{2}}_{\text{Estimated}} + \underbrace{\|x^{j}_{i} - \mu\|_{2}}_{\text{Clustering Error}}
\end{aligned}
\end{equation}
\begin{equation} \label{eq:10}
\begin{aligned}
    (\underbrace{\|q^{j} - x^{j}_{i}\|_{2}}_{\text{Actual}} - \underbrace{\|q^{j} - \mu\|_{2}}_{\text{Estimated}})^{2} \leq \underbrace{\|x^{j}_{i} - \mu\|^{2}_{2}}_{\text{Clustering Error}}
\end{aligned}
\end{equation}
Hence, through the above analysis, the argument is made that one only needs to minimize the squared clustering objective in \autoref{eq:3}, as it upper bounds the total distortion.

A natural question to ask is, why do existing PQ variants work well in the ID setting despite this upper-bound minimization approach? It is primarily because of two reasons. Firstly, as the graph traversal, initiated from a far off entry node, converges to a local region around an ID query, the first multiplicative term in \autoref{eq:3} progressively becomes smaller, to the extent that it no longer stays relevant. This is because the queries are in close proximity to their nearest base points in this scenario. Hence, the distortions progressively get smaller, and one only needs to focus on minimizing the second multiplicative term. Secondly, the difference in the true distances between a query and its $n$-th and ($n + 1$)-th near neighbors increases significantly, as $n$ gets smaller. This practically ensures that, even with distortions in the distance estimates, the search algorithm isn't easily confused between what appears to be near (based on a distorted estimate), and what is actually near (based on true distance) to a given query.

Extending the above argument, we can identify why existing PQ schemes fail to perform well in the OOD setting. The first multiplicative term in \autoref{eq:3} remains dominant for the entirety of the search process, since the query lies far away from any of the actual near points in the graph.

The average distance of a query to a base point is much larger in comparison to the average distance between two base points, or two queries. For high dimensional datasets, this means that for an OOD query, every base point appears to be roughly at the same distance from itself. Hence, the difference in the true distances between a query and its $n$-th and ($n + 1$)-th near neighbors is quite small for any value of $n$. Both of these phenomena, together, are able to trick the search algorithm into mistaking a far off point for a near neighbor and vice versa, leading to poor recall (\textbf{\autoref{fig:yandex-10m-pq-loss}}).

\subsection{Analysis for Inner Product Metric}
\label{sec:apq-ip-analysis}
For the inner product metric, following a similar approach, we have:
\begin{equation} \label{eq:11}
    \langle q^{j}, x^{j}_{i} \rangle = \langle q^{j}, x^{j}_{i} - \mu + \mu \rangle = \langle q^{j}, \mu \rangle + \langle q^{j}, x^{j}_{i} - \mu \rangle
\end{equation}
Thus, chunk-wise error between actual and estimated distance is:
\begin{equation} \label{eq:13}
\begin{aligned}
    \underbrace{\langle q^{j},  x^{j}_{i} \rangle}_{\text{Actual}} - \underbrace{\langle q^{j}, \mu \rangle}_{\text{Estimated}} = \langle q^{j}, x^{j}_{i} - \mu \rangle
\end{aligned}
\end{equation}
Hence~\autoref{eq:13} depicts the total distortion between the true inner product and the estimate distance. A positive value of the distortion term signifies that the base point appears less suitable to the query than it actually is, and correspondingly, a negative value of distortion term signifies that the base point appears more suitable. We can simplify the above as:
\begin{equation} \label{eq:14}
\begin{aligned}
    \langle q^{j},  x^{j}_{i} - \mu \rangle = \|q^{j}\| \|x^{j}_{i} - \mu\|\cos{\theta}
\end{aligned}
\end{equation}
where, $\theta$ is the angle between the two inner product vectors.

As before, we minimize the distortion between the true inner product and the estimate, by considering the following loss objective:

\begin{flalign} \label{eq:15}
    \mathcal{L}_{\text{APQ-IP}}(x^{j}_{i}, \mu, \mathcal{Q}_{x_{i}}) = \frac{1}{|\mathcal{Q}_{x_{i}}|} \sum_{q \in \mathcal{Q}_{x_{i}}}
        \begin{cases}
            \langle q,  x^{j}_{i} - \mu \rangle, & \text{if}\ x_{i} \in \text{H}_{T'}(q) \\
            |\langle q,  x^{j}_{i} - \mu \rangle|, & \text{otherwise}
        \end{cases} &&
\end{flalign}
where, $\mathcal{Q}_{x_{i}}$ and $\text{H}_{T'}(q)$ are obtained for the inner product metric as previously described.

\subsection{Limitations of APQ and AOPQ}
\label{sec:apq-pitfalls}

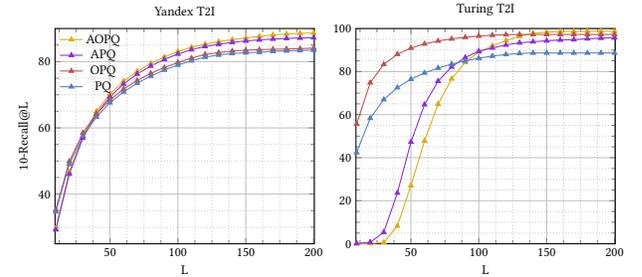
\begin{figure}[ht]
\flushleft
\begin{minipage}{\columnwidth}
    \begin{tikzpicture}[scale=0.5,font=\large]
    \begin{axis}[
        title=Yandex T2I,
        width=\linewidth,
        line width=0.8,
        grid=both,
        grid style={line width=.1pt, draw=gray!10},
        major grid style={line width=.2pt,draw=gray!50},
        minor grid style={line width=.3pt,draw=gray!100, dotted},
        minor tick num=3,
        tick label style={font=\large},
        legend style={nodes={scale=1.0, transform shape}},
        label style={font=\large},
        legend image post style={mark=triangle*},
        grid style={white},
        xlabel={L},
        ylabel={10-Recall@L},
        y tick label style={
            /pgf/number format/.cd,
            fixed,
            fixed zerofill,
            precision=0
        },
        legend style={at={(0,1), font=\normalsize}, anchor=north west, draw=none, fill=none},
        legend columns=1,
        ymin=25,
        ymax=90,
        xmin=9.5,
        xmax=200.5,
    ]
        \addplot[yyellow, mark=triangle*] coordinates
          {(10,29.67) (20,46.724) (30,57.679) (40,65.023) (50,70.22) (60,74.096) (70,77.073) (80,79.445) (90,81.417) (100,83.049) (110,84.349) (120,85.249) (130,86.049) (140,86.649) (150,87.149) (160,87.649) (170,88.049) (180,88.349) (190,88.549) (200,88.649)};
          \addlegendentry{AOPQ}

        \addplot[ppurple, mark=triangle*] coordinates
          {(10,29.255) (20,46.061) (30,56.92) (40,64.213) (50,69.41) (60,73.274) (70,76.276) (80,78.685) (90,80.662) (100,82.305) (110,83.614) (120,84.589) (130,85.244) (140,85.824) (150,86.241) (160,86.561) (170,86.809) (180,87.010) (190,87.120) (200,87.229)};
          \addlegendentry{APQ}

        \addplot[rred, mark=triangle*] coordinates
          {(10,35.256) (20,49.891) (30,58.456) (40,64.24) (50,68.463) (60,71.72) (70,74.315) (80,76.451) (90,78.22) (100,79.728) (110,81.062) (120,82.178) (130,82.703) (140,83.221) (150,83.456) (160,83.613) (170,83.724) (180,83.811) (190,83.913) (200,84.085)};
          \addlegendentry{OPQ}
          
        \addplot[bblue, mark=triangle*] coordinates
          {(10,34.672) (20,49.078) (30,57.545) (40,63.314) (50,67.617) (60,70.865) (70,73.48) (80,75.658) (90,77.46) (100,78.998) (110,80.262) (120,81.372) (130,81.998) (140,82.47) (150,82.678) (160,82.839) (170,83.133) (180,83.248) (190,83.351) (200,83.471)};
          \addlegendentry{PQ}
    \end{axis}
    \hspace{4.0cm}
    \begin{axis}[
        title=Turing T2I,
        width=\linewidth,
        line width=0.8,
        grid=both,
        grid style={line width=.1pt, draw=gray!10},
        major grid style={line width=.2pt,draw=gray!50},
        minor grid style={line width=.3pt,draw=gray!100, dotted},
        minor tick num=3,
        tick label style={font=\large},
        legend style={nodes={scale=1.0, transform shape}},
        label style={font=\large},
        legend image post style={mark=triangle*},
        grid style={white},
        xlabel={L},
        y tick label style={
            /pgf/number format/.cd,
            fixed,
            fixed zerofill,
            precision=0
        },
        legend style={at={(0,1), font=\normalsize}, anchor=north west, draw=none, fill=none},
        legend columns=1,
        ymin=0,
        ymax=100,
        xmin=9.5,
        xmax=200.5,
    ]
        \addplot[yyellow, mark=triangle*] coordinates
          {(10,0.023) (20,0.085) (30,0.585) (40,8.278) (50,27.066) (60,47.842) (70,64.854) (80,76.654) (90,84.271) (100,89.081) (110,91.988) (120,94.395) (130,96.395) (140,97.795) (150,98.294) (160,98.592) (170,98.653) (180,98.747) (190,98.755) (200,98.755)};

        \addplot[ppurple, mark=triangle*] coordinates
          {(10,0.148) (20,0.712) (30,5.326) (40,23.648) (50,47.282) (60,64.683) (70,75.547) (80,82.178) (90,86.525) (100,89.429) (110,90.99) (120,92.352) (130,93.353) (140,93.851) (150,94.274) (160,94.665) (170,94.853) (180,95.231) (190,95.536) (200,95.759)};

        \addplot[rred, mark=triangle*] coordinates
          {(10,55.638) (20,74.814) (30,83.349) (40,88.053) (50,90.924) (60,92.844) (70,94.199) (80,95.175) (90,95.917) (100,96.493) (110,96.903) (120,97.147) (130,97.225) (140,97.225) (150,97.225) (160,97.226) (170,97.226) (180,97.226) (190,97.226) (200,97.226)};
          
        \addplot[bblue, mark=triangle*] coordinates
          {(10,42.403) (20,58.31) (30,66.989) (40,72.559) (50,76.469) (60,79.382) (70,81.641) (80,83.459) (90,84.961) (100,86.207) (110,87.197) (120,87.931) (130,88.409) (140,88.631) (150,88.631) (160,88.631) (170,88.631) (180,88.631) (190,88.631) (200,88.631)};
    \end{axis}
\end{tikzpicture}
\end{minipage}
\caption{10-Recall@L scores for T2I datasets at 10M scale.}
\label{fig:combined-apq-recall-at-k}
\vspace{-5pt}
\end{figure}

At a high level, PQ and OPQ aim to minimize the quantization error for all base points with equal priority. Hence, this leads to large distortions on average, which are distributed more or less equally among the base points. APQ and AOPQ aim to minimize the quantization error over a chosen subset of base points, chalked out by their ``closeness'' to the query sample set. As such, while APQ and AOPQ lead to lower distortions for this chosen subset, they incur much higher distortions, for base points not considered to be ``close''. APQ and AOPQ are thus unsuitable for linear search routines not utilising re-ranking. Additionally, in the case of linear search with re-ranking, they also seem to underperform at low search budgets (\textbf{\autoref{fig:combined-apq-recall-at-k}}). This has no bearing for graph-based indices, as graph traversals are highly efficient and only look at a tiny fraction of the indexed dataset, which alleviates this problem.

\end{document}